%% file: main.tex
\documentclass{article}
\usepackage[T1]{fontenc}
\usepackage[hyperfootnotes=false]{hyperref}
\usepackage{microtype,graphicx,booktabs,placeins,multirow,xcolor}
\let\arxivaddcontentsline\addcontentsline
\usepackage[preprint]{icml2026}
\let\addcontentsline\arxivaddcontentsline
\usepackage{amsmath,amssymb,amsthm,mathtools,array,url,float,tabularx}
\newtheorem{proposition}{Proposition}

\newcommand{\E}{\mathbb E}

\newcommand{\ldm}{\textsc{ldm}}
\newcommand{\rlid}{RLiD}

\input{generated/macros}
\input{generated/visual_sokoban}
\makeatletter
\renewcommand{\ICML@preprint}{\textit{Preprint. October 21, 2026.}}
\renewcommand{\printAffiliationsAndNotice}[1]{%
 \global\icml@noticeprintedtrue
 {\let\thefootnote\relax\footnotetext{\textsuperscript{1}University of Illinois Urbana-Champaign.
 \textsuperscript{2}National University of Singapore.
 \textsuperscript{3}Zhejiang University.
 \textsuperscript{4}University of Pennsylvania.
 \textsuperscript{5}University of Rochester.\\[2pt]
 \Notice@String}}}
\makeatother
\hypersetup{pdftitle={RLiD: Reinforcement Learning in the Decision Plane for Language Agents and LLMs},pdfauthor={Ziyi Chen, Yan Zhang, Jianhui Wei, Jianing Yin, Daoan Zhang},pdfsubject={Research preprint}}
\icmltitlerunning{RLiD: Reinforcement Learning in the Decision Plane}
\begin{document}
\twocolumn[
\icmltitle{\texorpdfstring{RLiD: Reinforcement Learning in the Decision Plane\\for Language Agents and LLMs}{RLiD: Reinforcement Learning in the Decision Plane for Language Agents and LLMs}}
\begin{icmlauthorlist}
\icmlauthor{Ziyi Chen}{uiuc}
\icmlauthor{Yan Zhang}{nus}
\icmlauthor{Jianhui Wei}{zju}
\icmlauthor{Jianing Yin}{upenn}
\icmlauthor{Daoan Zhang}{rochester}
\end{icmlauthorlist}
\icmlaffiliation{uiuc}{University of Illinois Urbana-Champaign}
\icmlaffiliation{nus}{National University of Singapore}
\icmlaffiliation{zju}{Zhejiang University}
\icmlaffiliation{upenn}{University of Pennsylvania}
\icmlaffiliation{rochester}{University of Rochester}
\icmlkeywords{reinforcement learning, language agents, credit assignment, decision models, distillation}
\vskip 0.3in
]
\printAffiliationsAndNotice{}

\input{sections/abstract}
\input{sections/intro}
\input{sections/method}
\input{sections/setup}
\input{sections/results}
\input{sections/credit}
\input{sections/distill}
\input{sections/related}

\input{sections/conclusion}

\clearpage
\bibliography{refs}
\bibliographystyle{icml2026}

\newpage
\appendix
\onecolumn
\raggedbottom\setlength{\intextsep}{8pt plus 2pt minus 2pt}
\input{sections/appendix}

\end{document}

%% file: generated/macros.tex
\newcommand{\ArcTokQfifteen}{$75.9\pm0.7$ (3)}
\newcommand{\ArcTokQfifteenH}{1.17}
\newcommand{\ArcDelta}{2.6}
\newcommand{\ArcCost}{10}

\newcommand{\ArcTokQtwoMean}{85.8}
\newcommand{\ArcTokQtwoH}{0.76}
\newcommand{\ArcQtwoSentence}{on Qwen3.5-2B, whose rationale zero-shot is stronger, token GRPO is 1.8 points ahead at about four times the cost}
\newcommand{\RealVarAns}{71\%}

\newcommand{\RealVarAnsLen}{4}
\newcommand{\RealVarAnsInv}{0.4}
\newcommand{\RealVarCot}{97\%}

\newcommand{\RealVarCotLen}{43}
\newcommand{\RealVarCotInv}{84.6}
\newcommand{\RealVarConv}{With twice as many completions per cell (96) on a different random set of cells, the shares are 69\% and 98\%.}

\newcommand{\SokRows}{decision plane (\ldm{}) & $44.5\pm2.7$ & 2.3 & $44.9\pm3.3$\\
\ldm{} without letter regulariser & 5.9 & 2.9 & 6.2\\
token plane, answer only & $16.3\pm8.4$ & 2.9 & $17.2\pm4.5$\\
token plane, 30-word rationale & 5.9 & 5.6 & 6.2\\
token plane, rationale + decision & $14.5\pm1.1$ & 5.9 & $12.3\pm0.8$\\}

\newcommand{\AlfQtwoB}{$97.4\pm0.5$ / $98.5$}
\newcommand{\CreditLdmJ}{0.217}
\newcommand{\CreditLdmOracleCos}{0.88}

\newcommand{\CreditLdmOracleDJ}{$9.9\pm0.6$}
\newcommand{\CreditLdmGrpoCos}{0.54}
\newcommand{\CreditLdmGrpoSign}{0.52}
\newcommand{\CreditLdmGrpoDJ}{$5.7\pm0.7$}
\newcommand{\CreditLdmGigpoCos}{0.52}
\newcommand{\CreditLdmGigpoSign}{0.52}
\newcommand{\CreditLdmGigpoDJ}{$6.3\pm0.7$}
\newcommand{\CreditRecover}{58\%}
\newcommand{\CreditRecoverG}{64\%}
\newcommand{\CreditTokAnsLabelShare}{61\%--63\%}

\newcommand{\CreditTokAnsOracleDJ}{7.0 / 6.7}

\newcommand{\CreditTokAnsInv}{0.4}
\newcommand{\CreditTokCotLabelShare}{0.3\%--0.3\%}

\newcommand{\CreditTokCotOracleDJ}{4.0 / -0.2}
\newcommand{\CreditTokCotGrpoDJ}{2.9 / 0.0}
\newcommand{\CreditTokCotInv}{84.2}
\newcommand{\CreditTokRows}{\multicolumn{5}{@{}l}{\emph{Token, answer only} ($J{=}0.209$)}\\
\quad oracle & 0.67$^\dagger$ & 1.00 & $7.0\pm1.5$ / $6.7\pm1.7$ & 0.610 / 0.628\\
\quad GRPO & 0.18$^\dagger$ & 0.57 & $4.0\pm1.4$ / $4.3\pm1.3$ & 0.412 / 0.441\\
\quad GiGPO & 0.22$^\dagger$ & 0.57 & $5.2\pm1.1$ / $3.1\pm1.7$ & 0.442 / 0.495\\
\midrule
\multicolumn{5}{@{}l}{\emph{Token, 30-word rationale} ($J{=}0.196$)}\\
\quad oracle & 0.01$^\dagger$ & 1.00 & $4.0\pm1.6$ / $-0.2\pm2.0$ & 0.003 / 0.003\\
\quad GRPO & 0.06$^\dagger$ & 0.53 & $2.9\pm0.5$ / $0.0\pm2.7$ & 0.001 / 0.002\\
\quad GiGPO & 0.03$^\dagger$ & 0.53 & $2.9\pm3.2$ / $4.8\pm2.5$ & 0.002 / 0.002\\}
\newcommand{\CreditTokText}{In the token plane a large share of each update acts elsewhere: on fixed probe texts, only \CreditTokAnsLabelShare{} (answer only) and \CreditTokCotLabelShare{} (30-word rationale) of an update's absolute log-probability change falls on the answer-label token, and per-step improvements have smaller point estimates with greater estimation uncertainty; the rationale comparisons are inconclusive (Table~\ref{tab:credit}; Appendix~\ref{app:credit}). This is descriptive. The quantity Proposition~\ref{prop:rb} speaks about is measured next.

\paragraph{Realisation variance.} Proposition~\ref{prop:rb} predicts how much of the token-plane score is noise about how the action was written. For each of 24 FrozenLake cells we sample 48 completions and compute each completion's score $\nabla\log p(y\mid x)$. Grouping completions by outcome class (parsed label or invalid), the law of total variance splits the score's second moment into a decision part $\E\|\nabla\log\pi(c\mid x)\|^2$ and a realisation part $\E\|\nabla\log p(y\mid x,c)\|^2$. We estimate within-class variances with $n_c-1$ denominators and count classes with a single completion entirely as decision variance, which lowers the expected realisation numerator (not a guaranteed bound on each pooled ratio). \RealVarText}
\newcommand{\SokText}{The decision plane reaches $44.5\pm2.7\%$ (1.8--2.8 GPU-h per run; Figure~\ref{fig:curves}c), far above every token-plane variant. The answer-only token policy almost always parses but drifts towards label preferences (the top label takes 36--100\% of decisions vs.\ 25\% for a content-driven policy) and ends at 16.3\% on average. The 30-word rationale policy learns to parse within 50 iterations (invalid outputs fall from 86\% to about 1\%) yet stays at 6\% and ends at 5.9\%. The \ldm{} without its label regulariser collapses onto one label (5.9\%), so label collapse affects labelled menus in either plane; the decision plane's explicit label distribution makes it easy to correct. \RcdText}
\newcommand{\NoMemText}{a matched no-memory run with (observation, action) history reaches 91.8\%.}
\newcommand{\TokFailText}{Two failure modes are specific to the token plane: unparseable outputs (86\% of early Sokoban decisions with a short rationale) and credit spent on form (Section~\ref{sec:credit}).}

\newcommand{\LessonsTable}{On variable-size menus we use calibration (Eq.~\ref{eq:ldm}). Most menu sizes have few decisions per batch, so a per-size marginal regulariser can be satisfied most cheaply by flattening every row, i.e.\ it acts as an entropy bonus; we observed steadily rising entropy in ALFWorld pilots (logs not retained). On fixed four-action menus the regulariser is better. Episode-length weighting changes the surrogate so that long failed episodes, each with up to 50 negatively weighted decisions, do not dominate the update. With the environment's state key (which excludes the path-dependent memory), 63--95\% of ALFWorld decisions fall into anchor groups of size $\ge2$ (alfD\_q35\_2b\_h2).
\begin{table}[H]
\caption{Ablations behind the stability fixes of Section~\ref{sec:method} (test success, \%).}
\centering\small
\begin{tabularx}{\linewidth}{@{}>{\raggedright\arraybackslash}p{3.5cm}>{\raggedright\arraybackslash}X>{\raggedright\arraybackslash}p{2.8cm}>{\raggedright\arraybackslash}p{4.4cm}@{}}
\toprule
Issue & Setting & Without fix & With fix\\
\midrule
Too few updates per iteration & ALFWorld, Qwen3.5-2B, it.~40 & 27.6 (4 minibatches) & 45.5 (24 minibatches)\\
 & ALFWorld, Qwen2.5-1.5B, it.~40 & 8.2 (4) & 14.2 (24)\\
Letter bias on fixed 4-action menus & Sokoban, Qwen2.5-0.5B, final & 17.2 (calibration) & 37.9 / 55.3 / 42.2 (marginal regulariser, 3 seeds)\\
Greedy evaluation of a stochastic policy & ALFWorld, Qwen3.5-2B, no memory, final & 70.9 (greedy) & 77.6 (sampled, $T{=}1$)\\
\bottomrule
\end{tabularx}
\end{table}}
\newcommand{\RunIndexTable}{\begin{table}[H]
\caption{Run index (run names refer to logs released with the code). Runs on A100-40GB: arcX\_q7b, flX\_q3b, sok\_q05\_lr1e4, cd\_q15\_lr5e5, cda\_q15\_lr5e5; all others on H100.}
\label{tab:runs}
\centering\small
\begin{tabularx}{\linewidth}{@{}>{\raggedright\arraybackslash}p{3.9cm}>{\raggedright\arraybackslash}X@{}}
\toprule
Entry & Runs\\
\midrule
ARC \ldm{} Qwen2.5-1.5B / Qwen3.5-2B & arc\_q15\_lr1e5, arc\_q15\_s\{1,2\} / arc\_q35\_2b\_lr1e5, arc\_q35\_s\{1,2\}\\
ARC token GRPO (corrected parser) & tokrl\_arcfix\_q25\_1.5b\_s\{17,1,2\}, tokrl\_arcfix\_q35\_2b\_s\{17,1\}\\
Countdown plain / afterstate & cd\_q35\_2b\_m, cd\_q15\_lr5e5 / cda\_q15\_lr5e5, replicates cdaR\_q15\_s\{0..3\} (\texttt{--marg-mode micro --marg-rows adv}, our intended reconstruction of the original run's regulariser), cdaX\_q35\_\{2b,4b\}; tokrl\_cd\_*; token Countdown-after (aligned sampling) tokrl\_cdafix\_q25\_1.5b\_s\{17,1,2\} (superseded: tokrl\_cda\_q25\_1.5b); same-runner token arms mC\_\{ans,cot,rcd\}\_s\{0,1\} (cdaR hyperparameters plus \texttt{--policy}, \texttt{--fmt-pen 0}; the label regulariser acts only in the rcd path; mC\_rcd\_s1 run after the evaluation-RNG fix); \ldm{} without label regulariser cdaN\_q15\_s\{0,1\} (cdaR hyperparameters with \texttt{--marg 0}, \texttt{policy=ldm})\\
Countdown frozen protocol (Appendix~\ref{app:protocol}) & LR grid pr\_\{ldm,hyb,ans,cot\}\_lr\{2e5,5e5,1e4\}\_s0; seeds pr\_\{arm\}\_sel\_s\{1,2,3\}; stop-scored hybrid \texttt{--rcd-stop scored --rcd-early-stop 1}; exclusive timing tm\_*; sealed split se\_\{arm\}\_s\{0..3\}\_sh\{0,1\}; plain Countdown cdP\_q15\_s\{0,1\}, cdP\_q35\_2b\_s0, mP\_\{ans,rcd\}\_s0; earlier stop-corrected hybrid mCx\_rcd\_s\{0,1\}, mCx2\_rcd\_s\{2,3\}; strict re-evaluations strict\_mC\_*; aggregation \texttt{scripts/protocol\_report.py}\\
FrozenLake protocol & flP\_\{ldm,hyb\}\_lr\{1e5,1e4\}\_s0, flP\_\{ldm,hyb\}\_sel\_s\{1,2\}\\
Distill (Countdown-after) & teacher dump + BC cdD24k (results/bc/cdD24k), RL cdD24k\_rl; second student seed results/bc/cdD24k\_seed1, cdD24k\_rl\_s1; teacher-KL variant cdD24k\_T\\
Sokoban long / visual & sokL\_q05\_s\{0,1,2\} (1,000 it); vaS\_vl3b\_rl, vaD\_va\_q35\_4b\_to\_vl3b\_rl (eval\_t04\_\{best,final\}.jsonl)\\
Sokoban & sok\_q05\_lr1e4(\_s1,\_s2), sok\_q15\_m, sokM\_q35\_\{2b,4b\}; same-runner: mS\_ldm\_s\{0,1,2\}, mS\_ldmplain\_s\{0,1\}, token plane with aligned sampling mSf\_\{ans,cot,rcd\}\_s*\\
FrozenLake & fl\_q05\_m, fl\_q15\_m, flX\_q3b, flX\_q35\_2b\\
ALFWorld Qwen3.5 & alfD\_q35\_2b\_h2, alfE\_q35\_2b\_h2\_s1, alfD\_q35\_2b\_h2\_s2, alfD\_q35\_2b, alfE\_q35\_4b\_h2, alfm\_q35\_2b\_lr1e5, alfh6\_q35\_2b\_lr1e5\\
ALFWorld Qwen2.5 & alfE\_q15\_h2, alfH\_q15\_s1, alfH\_q7b; distill + RL: alfG\_q15\_bcrl (BC: bc\_alf\_q15), alfG\_q15\_bcrl\_s1 (BC seed 1: 133/134 unseen before RL), alfG\_q7b\_bcrl; all BC from the same teacher corpus (dumps/teacher\_alf\_q35\_2b\_train.jsonl, 15,224 decisions)\\
ALFWorld memory (matched) & alfD\_q35\_2b\_h2 vs.\ alfP\_q35\_2b\_nomem\\
Distill + RL & sokBC\_q15, sokBC\_q05r (0.5B; sokBC\_q05 crashed at it.~31), cdaBC\_q15, cdaE\_*\\
Credit audit / realisation variance & credit\_\{ldm,tok0,tok30\}\_q05\_base; realvar\_tok\{0,30\}(\_n96)\\
Standalone evaluations & results/evals: ALFWorld re-evaluation of alfD\_q35\_2b\_h2 and alfD\_q35\_2b final adapters (ldmrl/evaluate.py, shuffled menus, pre-action history 2 / 6, memory on, $T{=}1$ and greedy, 134 unseen + 140 seen games); BC-only evaluations of the distillation pipeline; Countdown on the token baseline's items\\
\bottomrule
\end{tabularx}
\end{table}
All \ldm{} and same-runner token-plane runs use LoRA rank 16 ($\alpha{=}32$), PPO clip 0.2, one PPO epoch, AdamW ($\beta{=}(0.9,0.999)$, no weight decay), gradient clipping 1.0, 16 tasks $\times$ 8 episodes per iteration (ARC: 64 $\times$ 8; Countdown: 32 $\times$ 8), and shuffled menus during training. Learning rates are $10^{-5}$ (Qwen3.5, ALFWorld), $3\cdot10^{-5}$--$5\cdot10^{-5}$ (Qwen2.5 on Countdown/ALFWorld) and $10^{-4}$ (0.5B Sokoban, Qwen3.5-4B Sokoban). Discount 0.95 for GiGPO; std-normalised advantages. Sokoban out-of-distribution maps were generated with a per-process randomised retry hash in all reported runs (about 7\% of OOD maps can differ between processes); the generator now uses a fixed CRC32 hash. Evaluation maps never use the retry path (four training IDs do). Splits are by seed, not by puzzle: 246 of the 512 Sokoban test seeds and 139 of the 256 validation seeds generate an initial puzzle identical to one in the training pool, so Sokoban results measure the shared distribution, not unseen-board generalisation (as for FrozenLake, Appendix~\ref{app:protocol}).}
\newcommand{\DistillAblTable}{\begin{table}[H]\caption{Countdown with afterstates: Qwen2.5-1.5B student, Qwen3.5-4B teacher. Test success after 0, 50 and 150 RL iterations (same 512-item subset, 32 tasks $\times$ 8 episodes per iteration), and final (1,024 items; RL iterations in brackets). Entropy: retained-update-batch decision entropy in nats at the first and last RL iteration (its sampled population changes with filtering and regularisation, so it is not a final-policy entropy).}\label{tab:distillabl}\centering\small\begin{tabular}{@{}lcccccc@{}}\toprule Student training & it.~0 & it.~50 & it.~150 & final & $H$ first & $H$ last\\\midrule pure RL & 11.3 & 30.7 & 49.0 & 60.4 (400) & 1.68 & 1.53\\ hard BC $\to$ RL & 51.8 & 56.2 & 54.7 & 56.2 (150) & 0.84 & 0.12\\ soft BC $\to$ RL & 53.5 & 56.6 & 55.1 & 56.5 (150) & 0.74 & 0.11\\ hard BC $\to$ RL + entropy floor & 51.8 & 56.4 & 55.7 & 55.9 (150) & 0.95 & 0.13\\ soft BC $\to$ RL + entropy floor & 53.5 & 56.8 & 55.5 & 57.0 (150) & 0.85 & 0.23\\\bottomrule\end{tabular}\end{table}}
\newcommand{\CreditDetails}{\textbf{Setting.} Qwen2.5-0.5B-Instruct with a freshly initialised LoRA adapter (rank 16), i.e.\ the start of RL. 16 FrozenLake maps (the first 16 test seeds), 4$\times$4, slip model with 80\% intended move, horizon 10, reward 1 on reaching the goal, $\gamma=1$. Prompts contain map, current cell and the fixed canonical menu (no history, memory or step counter), so the policy is Markov in (map, cell) and exact policy evaluation applies.
\textbf{Batches.} Each batch samples 8 episodes per map (about 730--890 decisions) with independent slip and action randomness. Advantages: GRPO (episode return minus group mean), GiGPO (plus centred return-to-go at (map, cell) anchors), and the exact $A^\pi_h(s,a)$.
\textbf{Update and $\Delta J$.} The batch gradient (ascent direction, no clipping, entropy or regularisers) is followed by an SGD step whose size is found by a line search (up to 6 trials, starting at unit-gradient scale 0.01) so that the mean KL between old and new decision distributions over visited cells is 0.01; achieved KLs (for the token plane, the smoothed Monte Carlo estimate used for step selection, not an independently verified semantic KL) average \AuditKLLdm{} (decision plane), \AuditKLAns{} (answer only) and \AuditKLCot{} (rationale). Weights are restored after each step. For the decision plane $\Delta J$ is exact. For the token plane the semantic policy is re-estimated from 32 sampled completions per cell before and after the step (paired seeds), and $\Delta J$ is computed by the same dynamic programming on the estimated policies.
\textbf{Token-plane reference and shares.} The token reference gradient is a Monte Carlo estimate from the same 32 completions per cell with exact DP advantages of the estimated policy; because policy, advantages and reference are coupled plug-in estimates, the token-plane cosines have no guaranteed bias direction. Shards share the reference sample and evaluation seed streams (which depend on the local batch index), so reported standard errors are conditional on that shared randomness. The decision share is the fraction of the summed absolute change in token log-probabilities, over 64 fixed probe completions, that falls on the answer-label token.
\textbf{Runs.} Decision plane: 32 batches, 166 GPU-seconds. Token, answer only: 8 batches, \AuditSecsAns{} GPU-seconds, invalid answers 0.5\%. Token, 30-word rationale: \AuditNCot{} batches from independent shards (different batch seeds, same policy and maps; shards merged batch-wise), invalid answers \CreditTokCotInv\% at the start of RL. Line-search bug note: an earlier version could select a null step when no trial came close to the KL target; it was fixed before all reported runs.

\textbf{Token-plane updates in detail.} In the token plane we describe where updates act. On fixed probe texts we measure how an update's absolute change in token log-probabilities is distributed over generated tokens. This is a descriptive statistic, not the realisation variance of Proposition~\ref{prop:rb}, which we measure directly below. For an answer-only policy, whose outputs are almost always ``\texttt{Answer: X}'' (\CreditTokAnsInv\% invalid), \CreditTokAnsLabelShare{} of the change falls on the answer label under oracle advantages; the rest moves format tokens. Its mean improvement per step is \CreditTokAnsOracleDJ$\times10^{-3}$ (whole-string / token-mean score), with \AuditVarRatioAns{} times the batch-to-batch variance in $\Delta J$ of the decision plane; that variance also includes the noise of estimating the token policy. With a 30-word rationale, \CreditTokCotInv\% of the base policy's outputs do not parse and only \CreditTokCotLabelShare{} of the change falls on the answer label. Improvements are inconclusive: \CreditTokCotOracleDJ$\times10^{-3}$ for oracle and \CreditTokCotGrpoDJ$\times10^{-3}$ for GRPO advantages, with oracle standard errors of \AuditSECot$\times10^{-3}$ (GRPO: \AuditSEGrpoCot$\times10^{-3}$). These rationale results rest on \AuditNCot{} batches from shards that share policy estimates. Token-plane cosines are not comparable with the decision plane's exact ones.

 For invalid token outputs, the oracle advantage of the randomly executed fallback action is used, so token-plane oracle updates retain fallback-action noise.}
\newcommand{\ValSelTable}{\begin{table}[H]\caption{Final vs.\ validation-selected adapter (test success, \%).}\centering\small\begin{tabular}{@{}lccc@{}}\toprule Run & final & val-selected & selected iteration\\\midrule ARC Qwen2.5-1.5B s1 & 78.4 & 78.4 & 100\\ ARC Qwen2.5-1.5B s2 & 79.1 & 78.8 & 90\\ ARC Qwen3.5-4B & 92.9 & 92.6 & 110\\ ARC Qwen3.5-9B & 94.6 & 93.9 & 20\\ Countdown-after Qwen2.5-1.5B & 60.4 & 60.4 & 400\\ Countdown-after Qwen3.5-4B & 67.4 & 67.9 & 325\\ Sokoban Qwen2.5-0.5B s1 & 55.3 & 55.3 & 300\\ Sokoban Qwen3.5-4B & 74.2 & 74.2 & 300\\ FrozenLake Qwen3.5-2B & 74.6 & 74.0 & 120\\ ALFWorld Qwen3.5-2B (T=1) & 97.8 & 96.3 & 140\\ ALFWorld Qwen3.5-2B s1 (T=1) & 97.0 & 97.0 & 150\\ ALFWorld Qwen3.5-4B (T=1) & 99.3 & 99.3 & 90\\ ALFWorld Qwen2.5-1.5B (T=1) & 79.9 & 77.6 & 130\\ ALFWorld Qwen2.5-1.5B s1 (T=1) & 85.8 & 85.8 & 150\\\bottomrule\end{tabular}\end{table}}
\newcommand{\ThroughputRows}{Qwen2.5-0.5B & decision plane (\ldm{}) & 1 (label) & 0.5 & 1.7 & 119.2 & --\\
 & token, answer only & 4.3 & 1.9 & 1.8 & 69.6 & 1.7$\times$\\
 & token, 30-word rationale & 45.9 & 8.3 & 2.3 & 24.2 & 4.9$\times$\\
\midrule
Qwen2.5-1.5B & decision plane (\ldm{}) & 1 (label) & 1.1 & 3.5 & 56.2 & --\\
 & token, answer only & 4.3 & 3.0 & 3.7 & 37.9 & 1.5$\times$\\
 & token, 30-word rationale & 20.7 & 10.9 & 4.5 & 16.6 & 3.4$\times$\\}
\newcommand{\SpeedAns}{1.5--1.7}
\newcommand{\SpeedCot}{3.4--4.9}

\newcommand{\BcSokQfive}{74.0}
\newcommand{\BcSokQfifteen}{70.3}
\newcommand{\BcCd}{53.5}
\newcommand{\BcAlf}{100}

\newcommand{\AlfSixShuf}{98.5 / 97.8}

\newcommand{\AuditNLdm}{32}
\newcommand{\AuditNAns}{8}
\newcommand{\AuditKLAns}{0.0099}
\newcommand{\AuditVarRatioAns}{1.4--1.8}

\newcommand{\AuditSecsAns}{3302}
\newcommand{\AuditNCot}{4}
\newcommand{\AuditKLCot}{0.0108}

\newcommand{\AuditSECot}{1.6 / 2.0}
\newcommand{\AuditSEGrpoCot}{0.5 / 2.7}

\newcommand{\AuditKLLdm}{0.0100}

\newcommand{\EstRows}{base, $R{=}2$ & 0.980 / 0.985 & 0.196 / 0.238 & $+0.06\pm0.34$ & 5/12\\
base, $R{=}3$ & 0.970 / 0.990 & 0.122 / 0.064 & $-0.23\pm0.48$ & 6/14\\
base, $R{=}4$ & 0.945 / 0.993 & 0.150 / 0.048 & $+0.57\pm0.36$ & 10/16\\
random init 1, $R{=}3$ & 0.848 / 0.912 & 0.319 / 0.209 & $+0.33\pm0.73$ & 8/16\\
random init 2, $R{=}3$ & 0.922 / 0.945 & 0.253 / 0.154 & $+0.59\pm0.73$ & 8/14\\
trained FrozenLake, $R{=}3$ & 0.956 / 0.975 & 0.181 / 0.137 & $+0.06\pm0.24$ & 8/15\\}
\newcommand{\EstCosWins}{6/6}
\newcommand{\EstErrWins}{5/6}
\newcommand{\EstN}{6}
\newcommand{\EstPairs}{87}
\newcommand{\EstMatchedRange}{12--16}
\newcommand{\EstKLRange}{0.0089--0.0104}
\newcommand{\EstPooled}{$+0.24\pm0.23$}
\newcommand{\EstClusters}{46}
\newcommand{\EstPooledPos}{45}
\newcommand{\EstPositive}{5}

\newcommand{\RvSokTrAnsContrib}{2}

\newcommand{\RvSokTrCot}{94}
\newcommand{\RvSokTrCotCi}{94 [91, 96]}

\newcommand{\RvSokTrCotValid}{99.7}

\newcommand{\RvCdaTrAnsContrib}{7}

\newcommand{\RvCdaTrCotCi}{9 [2, 27]}
\newcommand{\RvCdaTrCotRaw}{54}

\newcommand{\RvRows}{Sokoban 0.5B, base, answer only & 99.1 & 70 & 68 [57, 76] & 32 & 49\\
Sokoban 0.5B, trained, answer only & 100.0 & 0 & 0 [0, 0] & 2 & 0\\
Sokoban 0.5B, base, rationale & 16.1 & 94 & 53 [39, 67] & 32 & 58\\
Sokoban 0.5B, trained, rationale & 99.7 & 96 & 94 [91, 96] & 29 & 94\\
Countdown-after 1.5B, base, answer only & 99.5 & 29 & 30 [19, 41] & 30 & 18\\
Countdown-after 1.5B, trained, answer only & 93.7 & 2 & 0 [0, 0] & 7 & 0\\
Countdown-after 1.5B, base, rationale & 76.2 & 82 & 78 [69, 85] & 30 & 78\\
Countdown-after 1.5B, trained, rationale & 98.9 & 54 & 9 [2, 27] & 12 & 8\\}
\newcommand{\TrGrpoDthreeAuc}{$+0.17\pm0.14$}
\newcommand{\TrGrpoDthreeAucPos}{7/8}
\newcommand{\TrGrpoDthreeFinal}{$+0.009\pm0.004$}
\newcommand{\TrGrpoDthreeRel}{3}
\newcommand{\TrGrpoDthreeFinalCi}{$[-0.002, +0.019]$}
\newcommand{\TrGrpoDthreeAucCi}{$[-0.16, +0.50]$}

\newcommand{\TrOracleDthreeAuc}{$+0.16\pm0.05$}
\newcommand{\TrOracleDthreeAucPos}{4/4}
\newcommand{\TrOracleDthreeFinal}{$+0.015\pm0.008$}

\newcommand{\TrOracleDthreeAucCi}{$[-0.00, +0.32]$}
\newcommand{\TrRows}{GRPO adv., 3-token rationale & 8 & 0.291 / 0.300 & $+0.009\pm0.004$ (6) & $+0.17\pm0.14$ (7) & 214/240, 215/240\\
GRPO adv., 1-token rationale & 8 & 0.284 / 0.295 & $+0.011\pm0.009$ (4) & $+0.35\pm0.15$ (6) & 221/240, 214/240\\
oracle adv., 3-token rationale & 4 & 0.304 / 0.319 & $+0.015\pm0.008$ (3) & $+0.16\pm0.05$ (4) & 94/120, 107/120\\}

\newcommand{\DepthDthreeInitoneCos}{0.62 / 0.88}

\newcommand{\RatioDoneInitone}{1.3}
\newcommand{\RatioDthreeInitone}{7.1}
\newcommand{\RatioDthreeBase}{4.2}
\newcommand{\RatioDoneBase}{0.7}
\newcommand{\InvFourEightyNonfinite}{25}
\newcommand{\InvRows}{none & 0 & 8 & $+0.35$ $[-0.01, +0.71]$ (6) & $+0.011$ $[-0.011, +0.033]$ (4) & 10.0 / 10.1 & 221 / 214 & 0 / 0\\
4 tokens, low bias & 20 & 8 & $+0.50$ $[+0.07, +0.93]$ (6) & $+0.023$ $[-0.001, +0.046]$ (7) & 10.1 / 10.0 & 222 / 221 & 0 / 0\\
1 token, low bias & 25 & 8 & $+0.25$ $[-0.18, +0.68]$ (6) & $+0.016$ $[-0.010, +0.042]$ (5) & 10.1 / 10.0 & 203 / 214 & 0 / 0\\
4 tokens, high bias & 61 & 8 & $+0.83$ $[+0.25, +1.40]$ (6) & $+0.024$ $[-0.018, +0.067]$ (5) & 7.5 / 10.1 & 163 / 204 & 35 / 0\\
1 token, high bias & 64 & 8 & $+0.52$ $[-0.21, +1.25]$ (6) & $+0.019$ $[-0.004, +0.041]$ (6) & 10.2 / 10.0 & 203 / 200 & 0 / 0\\}
\newcommand{\InvPooled}{$+0.49\pm0.13$}
\newcommand{\InvPooledT}{3.8}
\newcommand{\InvPooledRel}{7}
\newcommand{\InvPooledSeeds}{7/8}
\newcommand{\InvPairs}{40}
\newcommand{\InvSlope}{$+0.38\pm0.56$}
\newcommand{\WkMWRel}{+1.3}
\newcommand{\WkMWCi}{$[-0.21, +0.34]$}
\newcommand{\WkMWPos}{5/8}
\newcommand{\WkWfourWRel}{+2.4}
\newcommand{\WkWfourWCi}{$[-0.14, +0.38]$}
\newcommand{\WkWfourWPos}{6/8}
\newcommand{\WkN}{8}

\newcommand{\RealVarText}{Pooled over cells, \RealVarAns{} of the answer-only policy's score second moment is realisation variance (\RealVarAnsLen{}-token outputs, \RealVarAnsInv\% invalid), and \RealVarCot{} with a 30-word rationale (\RealVarCotLen{} tokens, \RealVarCotInv\% invalid); \RealVarConv{} (The second run also re-initialises the LoRA adapter, so this shows stability across samples rather than paired convergence.) The decision plane has none. These are unweighted whole-string score moments within each token policy, relative to its own decision marginal; weighted by squared advantages, this part is exactly the token-plane estimator's excess variance (Proposition~\ref{prop:rb}). They are not training-time PPO variance, nor variance saved by a differently parameterised \ldm{}.}
\newcommand{\RcdText}{The format-valid \emph{rationale + decision} variant, with the same regulariser, ends at 14.5\%. Its loss is label PPO plus mean PPO over scored rationale tokens; unscored end-of-turn or malformed triggers fall back to a fixed suffix (62\% and 100\% of final decisions), so it does not fully credit when the rationale stops. Its rationale shrinks to one token in some runs and grows to the 64-token cap in others. With end-of-turn masked until the answer cue, it ends at 11.4\%: masking end-of-turn does not close the gap (a malformed answer cue still ends the rationale through the fixed suffix).}

%% file: generated/visual_sokoban.tex
\newcommand{\VlHoldoutNote}{5 of 5 frozen variant evaluations are complete. The pure-RL mean averages three training seeds; the matching seed-0 final is also shown. Each new variant uses seed 0. Means first average three decoding streams within each training seed. These final-checkpoint results differ from the validation-selected baseline results above. Holdout outcomes never select a recipe or checkpoint; Table~\ref{tab:published} remains the 128-puzzle benchmark.}
\newcommand{\VlHoldoutRows}{LDM, pure RL & $74.6\pm3.9$ \\
LDM, pure RL (seed 0) & $70.1$ \\
LDM + continuation control & $83.4$ \\
LDM + SPEAR replay & $84.0$ \\
\textbf{LDM + outcome curriculum} & $\mathbf{85.3}$ \\
LDM + SPEAR replay + curriculum & $84.4$ \\
LDM + outcome curriculum (from base) & $81.0$ \\}
\newcommand{\VlPublishedSpearFinal}{87.9}
\newcommand{\VlPureBudget}{200 it, 7.0--10.5 h}
\newcommand{\VlPureFinal}{84.8\pm3.0}
\newcommand{\VlPurePeak}{88.3}
\newcommand{\VlPurePeakBudget}{200 it, 10.5 h}
\newcommand{\VlPureTableFinal}{84.8\pm3.0}
\newcommand{\VlSpearAllRows}{LDM continuation control & $91.0$ & $89.8$ & 180+200 it, 7.0+4.3 h \\
LDM + SPEAR replay & $89.8$ & $91.9$ & 180+200 it, 7.0+4.6 h \\
LDM + outcome curriculum & $92.6$ & $90.6$ & 180+200 it, 7.0+4.7 h \\
LDM + replay + curriculum & $92.2$ & $92.7$ & 180+200 it, 7.0+5.3 h \\
LDM + replay, from base & $61.3$ & $55.5$ & 200 it, 7.4 h \\
LDM + curriculum, from base & $90.2$ & $89.1$ & 200 it, 5.7 h \\
LDM, unnormalized advantages & $82.4$ & $85.7$ & 200 it, 6.4 h \\
LDM + outcome curriculum (replicated) & $91.8\pm0.8$ & $91.1\pm0.8$ & 360--380 it, 11.7--15.6 h \\}

\newcommand{\VlSpearEvidenceNote}{The selected recipe averages three training seeds, including its selection pilot; other additions remain pilots.}
\newcommand{\VlSpearMainSelectionNote}{All completed LDM variants are shown; the verified replicated mean replaces its pilot result.}
\newcommand{\VlSpearNote}{New rows use final checkpoints and three decoding streams per training seed. Pilot recipes each use one training seed; verified replications use three independent training prefixes. Continued rows start from iteration 160 (prior search: 200 iterations, 10.5 GPU-h); iteration 180 (prior search: 200 iterations, 7.0 GPU-h); iteration 180 (prior search: 200 iterations, 10.0 GPU-h); budgets include those prior search costs. Every completed declared arm is reported in the main table, regardless of its benchmark score; the highest displayed visual-Sokoban mean is bold. Incomplete arms are ineligible. Recipes are selected by final-checkpoint greedy success on the 256 held-out-training selection puzzles (mean across declared training seeds; exact ties use frozen manifest arm order). Benchmark scores never choose a recipe. An arm appears once in the main table; its original pilot and verified replicated result are retained separately here. For independent-prefix replication, the recipe is frozen after all four continuation pilots finish, using their final selection success and manifest-order ties, including the control as a candidate. The main recipe row changes to the replicated mean only when all three declared training seeds are complete. Each seed starts from its own training-selected pure-RL checkpoint; hash and prior cost are verified separately. Where prefixes differ, budgets show the range of total policy-gradient iterations and total search GPU-hours per seed. Seed 0 participated in recipe selection, so its inclusion conditions the three-seed mean on that selection; seeds 1 and 2 are fresh replications. Fresh seeds 1--2 alone reach $91.3\pm0.9\%$ on the benchmark (mean $\pm$ s.d.\ over these two training seeds).}
\newcommand{\VlSpearRows}{\quad + continuation control & $89.8$ & 180+200 it, 7.0+4.3 h \\
\quad + SPEAR replay & $91.9$ & 180+200 it, 7.0+4.6 h \\
\quad + outcome curriculum & $91.1\pm0.8$ & 360--380 it, 11.7--15.6 h \\
\textbf{\quad + SPEAR replay + curriculum} & $\mathbf{92.7}$ & 180+200 it, 7.0+5.3 h \\
\quad + SPEAR replay (from base) & $55.5$ & 200 it, 7.4 h \\
\quad + outcome curriculum (from base) & $89.1$ & 200 it, 5.7 h \\
\quad + unnormalized advantages & $85.7$ & 200 it, 6.4 h \\}

\newcommand{\VlSpearTopMethod}{LDM + SPEAR replay + curriculum}
\newcommand{\VlSpearTopResult}{92.7}
\newcommand{\VlTeacherBudget}{200 it, 4.4 h}
\newcommand{\VlTeacherFinal}{87.5}

%% file: sections/abstract.tex
\begin{abstract}
Reinforcement learning (RL) has improved the capabilities of large language models (LLMs) across a wide range of tasks. However, token-level RL assigns an action's advantage to the entire generated response, including tokens specifying the parsed action, reasoning chains, and malformed content. Although the environment executes only the parsed action, variation in how the action is expressed in natural language can introduce additional variance into policy-gradient estimates. We propose \textbf{\rlid{}} (\textbf{R}einforcement \textbf{L}earning \textbf{i}n the \textbf{D}ecision plane), a framework that directly optimizes the probabilities of executable action options. \rlid{} compiles observations, an optional compact memory, and a labeled action menu into the model input, and uses a large decision model (\ldm{}) to select an action in a single forward pass. By eliminating autoregressive token generation and bypassing action parsing, \rlid{} avoids invalid action outputs and improves RL rollout efficiency. Our analysis shows that, within a fixed text policy, marginalizing over alternative realizations of the same action removes a zero-mean gradient component without changing the expected update. We further distill trained \ldm{} policies into generative LLMs through supervision on their action trajectories. Experiments on ARC, Countdown, and Sokoban demonstrate improved performance or lower computational cost across several settings. On ALFWorld, the complete training and distillation pipeline reaches 99.3\% success in 12.7 H100 GPU-hours, compared with 86.6\% in 17.3 GPU-hours for token-level RL of the same LLM. With compiled memory, our strongest Qwen2.5-1.5B \ldm{} and distilled generative LLM achieve 100\% success on all 134 unseen and 140 seen ALFWorld evaluation games.
\end{abstract}

%% file: sections/intro.tex
\section{Introduction}
\label{sec:intro}
\begin{figure*}[t]
\centering
\includegraphics[width=\textwidth]{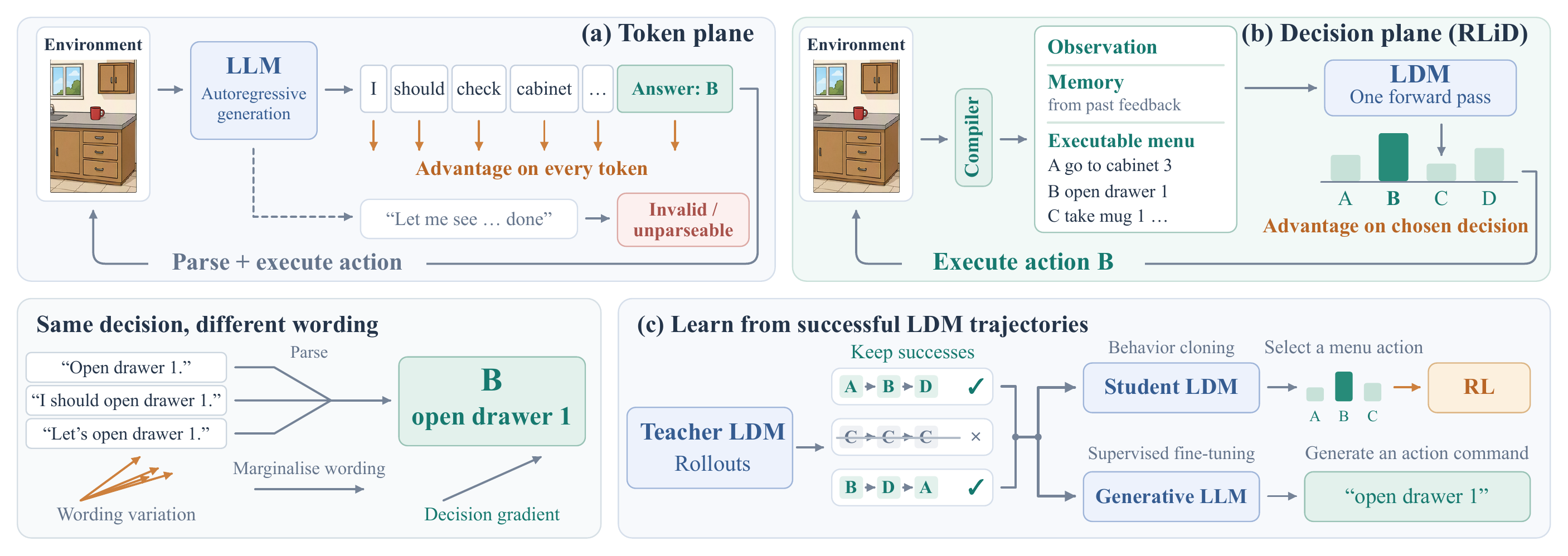}
\caption{\textbf{(a)} token-level RL credits every generated token, can produce unparseable outputs and pays for generation. \textbf{(b)} \rlid{}: a large decision model reads the compiled state and labelled menu in one pass. The advantage moves the probability of the executed decision directly (Proposition~\ref{prop:rb}). \textbf{(c)} successful teacher trajectories supervise student LDMs through behavior cloning followed by decision-level RL, or generative LLMs through supervised fine-tuning on action commands.}
\label{fig:overview}
\end{figure*}

Reinforcement learning (RL) has improved the capabilities of large language models (LLMs) in reasoning and interactive decision-making tasks \citep{shao2024deepseekmath,wang2025ragen,feng2025gigpo}. For language agents, RL typically operates in the \emph{token plane}. The model generates a response, often a rationale followed by an action command, and the action's advantage is assigned to every generated token.

However, the environment executes only the parsed action, while training also updates reasoning chains, formatting tokens, and potentially malformed content. Different natural-language responses can specify the same action yet produce different policy-gradient estimates. This variation can introduce additional gradient variance without changing the executed decision. Autoregressive generation also adds rollout cost, and unparseable outputs can waste interactions. These observations motivate directly optimizing the probabilities of executable action options.

We propose \rlid{} (Reinforcement Learning in the Decision Plane) for tasks with a finite menu of available actions, such as admissible commands, grid-world moves, or answer options. A compiler prepares the observation, an optional compact memory from past feedback, and a labeled action menu. A large decision model (\ldm{}) reads this input in one forward pass and outputs a categorical distribution over menu labels \citep{chen2026ldm}. Existing RL objectives then optimize the selected actions' probabilities. This formulation eliminates autoregressive token generation during rollouts and bypasses action parsing, ensuring that every selected action belongs to the executable menu (Figure~\ref{fig:overview}).

Our analysis relates text-level and decision-level credit. When reward depends on text only through the parsed action, the whole-string score decomposes into the induced action score and a zero-mean realization score. Within the same text policy, marginalizing over alternative expressions preserves the expected update and removes the associated variance component (Proposition~\ref{prop:rb}). An \ldm{} models actions directly and has no text-realization component. Temporal credit assignment still requires estimators such as GiGPO \citep{feng2025gigpo}. The analysis does not guarantee superiority over generative policies, whose reasoning tokens provide additional computation.

Successful \ldm{} trajectories can also supervise a generative LLM to write action commands as free text. This provides a post-training route that learns decisions with RL and then transfers them to text generation. We evaluate performance and computational cost on ARC, Countdown, Sokoban, and ALFWorld, including both decision policies and generative students. Because menus and memory affect the available information, we compare identical menus where possible and report interface differences and settings where the method does not help.

\paragraph{Contributions.}
\begin{itemize}
\item \textbf{\rlid{}.} Single-pass language policies over compiled menus, their Rao--Blackwell relation to token-level RL, and simple fixes for four failure modes (Section~\ref{sec:method}).
\item \textbf{Matched comparisons} (Section~\ref{sec:results}). ARC: +\ArcDelta{} points at about $\ArcCost\times$ lower cost with Qwen2.5-1.5B, while \ArcQtwoSentence. Countdown-after (frozen protocol): significant +12 to +14 points over token RL, mostly from the label regulariser. Performance does not differ significantly from a rationale-plus-label hybrid, while training takes 0.46$\times$ as long. Plain Countdown is weak at 1.5B. Same-runner Sokoban: 44.5\% vs.\ at most 26\%.
\item \textbf{An exact credit audit} (Section~\ref{sec:credit}). On FrozenLake, exact advantages give a batch gradient with cosine \CreditLdmOracleCos{} to the true gradient. Realisation noise accounts for \RealVarAns{} (answer only) and \RealVarCot{} (30-word rationale) of a token policy's unweighted score second moment. The decision plane has no realisation noise.
\item \textbf{Distillation across models} (Sections~\ref{sec:results}--\ref{sec:distill}). Decision-plane RL of a teacher, distillation on the shared menu and student RL give 100\% unseen and seen success for Qwen2.5-1.5B and 7B with 2.1--5.1 GPU-h of student RL, above state-of-the-art token RL (GiGPO 86.7 / 90.8, DARS 96.9 / 98.6). On Countdown-after the distilled student exceeds its teacher (70.2 vs.\ 67.4).
\item \textbf{Back to the generative LLM} (Section~\ref{sec:l2l}). Sequence-level distillation gives a free-text Qwen2.5-1.5B 100\% unseen and seen ALFWorld success with compiled memory, greedily and under sampling. Success reaches 99.3\% without the admissible-action list and 90.5\% without memory (GiGPO: 86.7\%). In a separately timed end-to-end run, the whole route reaches 99.3\% success in 12.7 GPU-h, whereas token RL of the same LLM needs 17.3 GPU-h to reach 86.6\%.
\end{itemize}

%% file: sections/method.tex
\section{\rlid{}: RL in the Decision Plane}
\label{sec:method}

\paragraph{Compiled decision problems.}
An agent faces a partially observed task. A \emph{compiler} $C$ maps each interaction state to a finite menu of executable actions $M_t=(m_{t,1},\dots,m_{t,k_t})$, an observation rendering $o_t$, and optionally a compact \emph{memory} $b_t$ (a belief state computed from past feedback, never from privileged simulator information). The compiler is ordinary code: benchmark-native admissible actions (ALFWorld, Sokoban, FrozenLake), choices of an answer key (ARC), or a menu written once by a frontier language model (Countdown, Section~\ref{sec:setup}). Menus may also show the \emph{afterstate} of each action, i.e., its deterministic effect. All of these change the decision problem the policy faces. We therefore report compiled menus as part of the method and compare policies on identical menus, except where an interface difference is marked.

\paragraph{Single-pass decision models.}
A \emph{large decision model} (\ldm{}) \citep{chen2026ldm} is a language-model backbone $f_\theta$ read out once per decision. The prompt $x_t$ contains the task, a short history of (observation, action) pairs, $b_t$, $o_t$ and the menu with entry $j$ prefixed by the single-token label $\ell_j\in\{\texttt{A},\dots,\texttt{Z},\texttt{a},\dots,\texttt{z},\texttt{0},\dots,\texttt{9}\}$. With $h_\theta(x)$ the final hidden state at the last prompt position and $W$ the output embedding,
\begin{equation}
\pi_\theta(j\mid x)=\operatorname{softmax}_{j\le k}\big(z_j\big),\quad z_j=W_{\ell_j}^{\top}h_\theta(x)-\beta_{k,j}.
\label{eq:ldm}
\end{equation}
$\beta_{k,\cdot}$ is a content-independent letter calibration, described below. A decision costs one prefill and no generation. The policy is an explicit categorical distribution over executable actions: there are no invalid outputs, and no length normalisation is needed for multi-token action strings, unlike scoring every action by its token likelihood \citep{tan2024twosome,carta2023glam}.

\paragraph{Decision-level policy optimisation.}
We collect $G$ episodes per task with the current policy. With episode returns $R^{(i)}$, the GRPO-style advantage \citep{shao2024deepseekmath} is $\hat A^{(i)}_{\text{ep}}=(R^{(i)}-\bar R)/(\sigma_R+\epsilon)$. GiGPO \citep{feng2025gigpo} adds a step-level term that groups decisions taken from the same anchor state (same task and same compiled observation) and centres their discounted returns-to-go. Both estimators act on decisions, so they apply unchanged to an \ldm{}. Every decision $(x_t,j_t)$ contributes the PPO-clip term \citep{schulman2017ppo}
\begin{equation}
\ell_t(\theta)=-\min\!\big(\rho_t\hat A_t,\ \operatorname{clip}(\rho_t,1\pm\epsilon)\hat A_t\big),\; \rho_t=\tfrac{\pi_\theta(j_t\mid x_t)}{\pi_{\text{old}}(j_t\mid x_t)}.
\end{equation}
We train LoRA adapters \citep{hu2022lora} only.

\paragraph{What the decision plane changes.}
Contrast this with the usual \emph{token plane}. There the policy writes a string $y\sim p_\theta(\cdot\mid x)$, typically a short rationale followed by an answer, and the environment executes the parsed action $a=g(y)$, or treats $y$ as invalid. Token-level REINFORCE/PPO uses the score of the whole string.
\begin{proposition}[Decision-plane credit is the Rao--Blackwellised token credit]
\label{prop:rb}
Let the reward depend on $y$ only through $a=g(y)$, and let $\pi_\theta(a\mid x)=\sum_{y:g(y)=a}p_\theta(y\mid x)$ be the induced decision policy. Then
\[
\nabla\log p_\theta(y\mid x)=\underbrace{\nabla\log\pi_\theta(a\mid x)}_{\text{decision score}}+\underbrace{\nabla\log p_\theta(y\mid x,a)}_{\text{realisation score}},
\]
and $\E[\nabla\log p_\theta(y\mid x,a)\mid x,a]=0$. Hence for any advantage $\hat A$ that is conditionally independent of $y$ given $(x,a)$, $\E[\hat A\,\nabla\log p_\theta(y\mid x)\mid x,a,\hat A]=\hat A\,\nabla\log\pi_\theta(a\mid x)$, and
\begin{align*}
&\operatorname{tr}\operatorname{Cov}\big[\hat A\,\nabla\log p_\theta(y\mid x)\big]-\operatorname{tr}\operatorname{Cov}\big[\hat A\,\nabla\log\pi_\theta(a\mid x)\big]\\
&\qquad=\E\big[\hat A^{2}\,\|\nabla\log p_\theta(y\mid x,a)\|^{2}\big]\;\ge\;0.
\end{align*}
\end{proposition}
The proof uses the chain rule and total covariance (Appendix~\ref{app:proof}). This is an identity \emph{within one policy}: marginalising its text removes realisation noise without changing the expected update. An \ldm{} parameterises the action distribution directly, so this term is zero. The proposition does \emph{not} compare differently parameterised policies, nor whole-string scores with our token-averaged PPO baseline; rationales also add computation. Sections~\ref{sec:results} and \ref{sec:credit} test these empirical differences.

The proposition concerns \emph{where} credit lands, not \emph{when}. Temporal credit assignment, i.e., which decisions in an episode caused the outcome, is still estimated by $\hat A$, with all the usual noise. The credit audit (Section~\ref{sec:credit}) measures both against exact dynamic-programming advantages.

\paragraph{Making the decision plane trainable.}
Four issues recur across environments, each with a simple decision-level fix (Appendix~\ref{app:lessons}). \emph{(i) Label bias:} pretrained models prefer some labels regardless of content \citep{zheng2024mcq}, and RL can amplify this into collapse onto one label. We shuffle menus during training. We either add a penalty $\mathrm{KL}(u\,\|\,\bar\pi_k)$ between the uniform distribution and the mean label distribution of each micro-batch, per menu size (Sokoban, Qwen2.5 runs on Countdown and FrozenLake), or subtract a calibration $\beta_k$ that receives no gradient and is updated after each iteration from the observed label marginal (ALFWorld, Qwen3.5 runs on Countdown and FrozenLake); the penalty works better on small fixed menus, the calibration on menus whose size varies (Appendix~\ref{app:lessons}). \emph{(ii) Length weighting} (ALFWorld; Qwen3.5 Countdown runs): each decision is weighted by $\bar T/T^{(i)}$, changing the surrogate so that every episode carries equal total weight. \emph{(iii) Anchors:} the GiGPO anchor key excludes path-dependent memory, which would make almost every state unique. \emph{(iv) Evaluation:} greedy decoding can loop, so we also report sampled ($T{=}1$) evaluation.

\paragraph{Distill + RL.}
Teacher choices in successful rollouts supervise another \ldm{} on freshly shuffled menus, followed by decision-level RL. They also supervise a generative LLM on action strings, transferring the policy back to the token plane (Section~\ref{sec:distill}).

%% file: sections/setup.tex
\section{Experimental Setup}
\label{sec:setup}

\paragraph{Environments.}
We use five environments with different credit-assignment structure (Appendix Table~\ref{tab:envs}).
\emph{ARC-Challenge} \citep{clark2018arc} is a one-step bandit: 1,119 training, 299 validation and 1,172 test questions, with the answer options as the menu.
\emph{Countdown} \citep{pan2025tinyzero} requires combining 3--4 numbers with arithmetic to hit a target in up to three reductions. Its menu of ``slot $i$ op slot $j$'' reductions was written once by a frontier model (gpt-6-astra) from a task description and audited. The afterstate variant additionally shows each reduction's result and the remaining numbers. We use 100k training and 1,024 held-out problems.
\emph{Sokoban} (6$\times$6, one box) and \emph{FrozenLake} (4$\times$4, slippery; slips are seeded per task, so the episodes of a group share slip randomness, unlike the credit audit) follow the RAGEN protocol \citep{wang2025ragen}: at most 10 actions, four moves, 512 test seeds and 512 out-of-distribution seeds (Sokoban 8$\times$8 with two boxes; FrozenLake 6$\times$6).
\emph{ALFWorld} \citep{shridhar2021alfworld} has 50-step household tasks with admissible commands as the menu. We follow the GiGPO protocol \citep{feng2025gigpo}: 134 unseen and 140 seen evaluation games. Its compiled memory tracks location, held object, contents of visited receptacles, unvisited receptacles and completed operations, computed from textual feedback only.

\paragraph{Models and training.}
Backbones are Qwen2.5-Instruct (0.5B, 1.5B, 3B, 7B) \citep{qwen2025qwen25} and Qwen3.5 (2B, 4B, 9B) \citep{qwen2026qwen35}, each with LoRA rank 16 on all linear layers of the language model. An iteration samples 16 tasks with $G{=}8$ episodes each (ARC: 64 questions; Countdown-after: 32 problems), so groups are formed per task. Sequential environments use GiGPO advantages, ARC uses GRPO. Updates are PPO-clip(0.2) with AdamW and four to 24 minibatches per iteration. Learning rates (1e-5 to 1e-4) were chosen per backbone on validation data. Checkpoints are selected on a validation split: ARC's own, otherwise 256 held-out training tasks. The test split is also logged every 10--25 iterations for learning curves, which permits adaptive peeking during development. We therefore report the \emph{final} checkpoint (Countdown-after: a common matched budget), not the best test point, and list validation-selected results in the appendix. Appendix~\ref{app:hparams} lists every run.

\paragraph{Token-plane baselines.}
(a) \emph{Own runs.} On ARC and plain Countdown: GRPO with a short rationale (TRL); on Countdown-after: our token-averaged policy gradient on the same menus, and token arms inside the \ldm{} runner. On Sokoban the token plane runs \emph{inside the same runner} (environment, menus, history, grouping, GiGPO, PPO, optimiser, learning rate, LoRA, interactions): the policy writes text ending in ``\texttt{Answer: <label>}'', every generated token receives the decision's advantage (token-averaged per decision), and an unparseable answer executes a random action with a 0.1 penalty. Variants: answer only, a rationale of at most 30 words, and a rationale followed by a constrained label readout with the \ldm{}'s label regulariser (``rationale + decision''); we also report the \ldm{} without its regulariser. The credit audit uses its own sampler and plain policy-gradient steps for both planes. (b) \emph{Published results}: RAGEN (StarPO) on Sokoban and GiGPO/GRPO on ALFWorld, for context only, since implementations, prompts, evaluation temperatures and hardware differ.

\paragraph{Compute accounting.}
Every run logs wall-clock seconds per phase (rollout, update, evaluation) and token counts. GPU-hours (GPU-h) are the number of GPUs times wall clock, including periodic evaluation, on H100 unless marked A100. Our runs share a node with other jobs, which inflates wall clock for both planes; same-runner comparisons are therefore also reported per training interaction. For published baselines we quote the authors' configurations; where they do not report compute we give an estimate and mark it as such.

%% file: sections/results.tex
\section{Decision Plane versus Token Plane}
\label{sec:results}

\begin{figure*}[t]
\centering
\includegraphics[width=\textwidth]{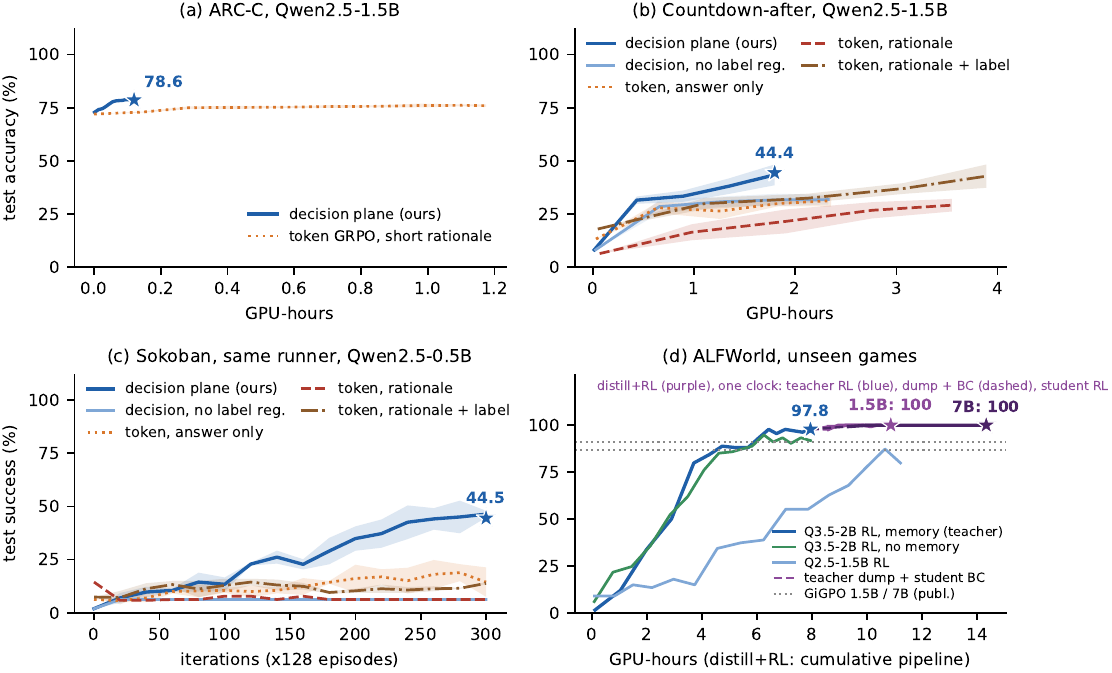}
\caption{Test success during training. Solid lines: decision plane (ours); dotted/dashed: token plane and published baselines. Stars mark our main runs' final evaluations (seed mean on the full test set, as reported in the tables); curves use periodic evaluations on test subsets. In (a)--(c) each line is the mean over seeds at the same iteration (three seeds per plane in (a), eight per arm in (b), completed seeds in (c)) and the band is $\pm$ one s.d.; the GPU-h axis is the seed mean. (a) ARC-C. (b) Countdown with afterstate menus, frozen protocol (GPU-h on shared H100s; exclusive timing in Table~\ref{tab:matched}). (c) Sokoban with the same runner. (d) ALFWorld, one run per curve, on one pipeline clock: the distill + RL route (purple) continues from the teacher's RL (blue, 7.9 h) through the teacher dump and student BC (dashed; on an exclusive H100 the dump takes 0.17 h and BC 0.45 h for the 1.5B student and 1.15 h for the 7B student) into student RL (solid). BC alone already solves 99.3--100\% of unseen games (Section~\ref{sec:distill}).}
\label{fig:curves}
\end{figure*}

\subsection{Matched comparisons}
Table~\ref{tab:matched} compares the two planes on the same tasks, rewards and backbones, on identical menus except where marked.

\begin{table*}[t]
\caption{Decision plane (\ldm{}) vs.\ token-level RL: test success (\%), mean $\pm$ s.d.\ over seeds (seed counts and per-seed results in Appendix~\ref{app:protocol}). Bold: larger paired mean, not significance. Episodes and H100 GPU-h through the iteration (incl.\ periodic evaluation; TRL timers exclude the initial zero-shot evaluation; Countdown-after: see $^{c}$): decision / token. Countdown menu-policy pairs share one runner (environment, menus, grouping, advantages, PPO, optimiser, episodes; learning rate tuned per arm); greedy evaluation.}
\label{tab:matched}
\centering\footnotesize
\setlength{\tabcolsep}{3pt}
\begin{tabular}{@{}llcccc@{}}
\toprule
Task, backbone & Token-plane comparator & Episodes (k) & Decision & Token & GPU-h\\
\midrule
\multicolumn{6}{@{}l}{\emph{One-step decisions (ARC-C, all 1,172 test questions)}}\\
Qwen2.5-1.5B & GRPO (TRL), short rationale & 61.4 / 26.7 & {\boldmath\bfseries $78.6\pm0.5$} & $75.9\pm0.7$ & 0.12 / \ArcTokQfifteenH\\
Qwen3.5-2B & GRPO (TRL), $\le$20-word rationale & 61.4 / 17.8 & $84.0\pm0.4$ & {\boldmath\bfseries $85.8\pm0.7$} & 0.15--0.18 / \ArcTokQtwoH\\
\midrule
\multicolumn{6}{@{}l}{\emph{Countdown-after, Qwen2.5-1.5B, frozen protocol$^{c}$: test (1,024) / untouched problems (2,048, mean)}}\\
\ldm{} & answer only & 25.6 & {\boldmath\bfseries $44.4\pm5.2$ / 42.2} & $32.0\pm2.2$ / 30.3 & 0.71 / 1.06\\
\ldm{} & 30-word rationale & 25.6 & {\boldmath\bfseries $44.4\pm5.2$ / 42.2} & $30.7\pm2.5$ / 28.4 & 0.71 / 1.53\\
\ldm{} & rationale + legal label$^{a}$ & 25.6 & {\boldmath\bfseries $44.4\pm5.2$ / 42.2} & $44.2\pm5.5$ / 41.9 & 0.71 / 1.57\\
\ldm{}, no label reg. & answer only & 25.6 & {\boldmath\bfseries $33.1\pm2.0$ / 31.6} & $32.0\pm2.2$ / 30.3 & 0.70 / 1.06\\
\bottomrule
\end{tabular}
\par\smallskip\begin{minipage}{\textwidth}\footnotesize
$^{a}$Hybrid: a rationale, then a menu-label readout with the \ldm{}'s label regulariser; end-of-turn is masked and a malformed answer cue ends the rationale with its trigger token scored, so every sampled stop is scored. $^{c}$Learning rate per arm chosen on validation, then several seeds; final checkpoints, greedy, strict success (unparseable text fails); untouched problems are in neither training nor test data; GPU-h: per-iteration rollout + update time on one exclusive H100 $\times\,100$, no evaluation (elapsed hours through the last training iteration, incl.\ periodic evaluation, in Appendix~\ref{app:protocol}). Plain Countdown (no afterstates) is weak for both planes (Appendix Table~\ref{tab:plaincd}); with distillation the Countdown-after student reaches 70.2\% (Table~\ref{tab:distill}).
\end{minipage}
\end{table*}

\paragraph{One-step decisions.}
On ARC the \ldm{} improves Qwen2.5-1.5B from 73.0\% to $78.6\pm0.5\%$ in 0.12 GPU-h; token-level GRPO with a short rationale reaches \ArcTokQfifteen{} in \ArcTokQfifteenH{} GPU-h (corrected reruns; an earlier version could not parse ARC's numeric labels). On Qwen3.5-2B, whose rationale zero-shot accuracy exceeds its single-pass accuracy, the planes end at 84.0 vs.\ \ArcTokQtwoMean{} at a quarter of the cost or less.

\paragraph{Planning with a compiled menu.}
On plain slot-reference Countdown menus both planes are weak (Appendix Table~\ref{tab:plaincd}); compiling \emph{afterstates} (the numbers left after each reduction) into the menu helps both. Under a frozen protocol (validation-tuned learning rate per arm, final checkpoints, strict success; Appendix~\ref{app:protocol}) the \ldm{} reaches 44.4\% on test and 42.2\% on 2,048 untouched problems. It beats answer-only and rationale token RL by 12--14 points (statistically significant, paired $p<0.001$) and shows no significant difference from the scored hybrid at 26 vs.\ 56\,s per iteration (Appendix~\ref{app:case}). Most of this lead comes from the label regulariser, which a menu policy can apply directly: without it the \ldm{} is level with answer-only token RL (no significant difference), at lower cost. Distillation lifts it to 70.2\% (Section~\ref{sec:distill}).

\paragraph{Same runner, multi-step.}
Table~\ref{tab:sokoban} compares the planes inside one runner on Sokoban (Qwen2.5-0.5B, 300 iterations of 128 episodes, learning rate $10^{-4}$, GiGPO). Environment, two-step history, menus, rollout grouping, advantages, PPO, optimiser and LoRA are shared code. \SokText

\begin{table}[t]
\caption{Same-runner Sokoban (Qwen2.5-0.5B, 512 test seeds, greedy). Final: test success after 300 iterations, mean $\pm$ s.d.\ over completed seeds. GPU-h per run (shared node); last column: success at 2.5 GPU-h (curves, 256 seeds).}
\label{tab:sokoban}
\centering\small
\resizebox{\linewidth}{!}{\begin{tabular}{@{}lccc@{}}
\toprule
Policy & final & GPU-h & at 2.5 GPU-h\\
\midrule
\SokRows
\bottomrule
\end{tabular}}
\end{table}

\paragraph{Cost per decision.}
Timing the same 256 Sokoban decisions per policy on one GPU (rollout plus training pass; Appendix Table~\ref{tab:throughput}), the decision plane processes \SpeedAns$\times$ more decisions per second than an answer-only token policy and \SpeedCot$\times$ more than one with a 30-word rationale: generation scales with output length, a decision costs one prefill.

\subsection{Against published multi-step agents}
Table~\ref{tab:published} gives contextual comparisons; protocols, prompts and hardware differ. Additional ALFWorld backbones are in Appendix Table~\ref{tab:alf-additional-backbones}.

\begin{table}[t]
\caption{Published comparisons (context only): success (\%) and per-run budget. LDM visual rows use final checkpoints.}
\label{tab:published}
\centering\scriptsize
\setlength{\tabcolsep}{2.5pt}
\renewcommand{\arraystretch}{0.9}
\begin{tabularx}{\columnwidth}{@{}>{\raggedright\arraybackslash}Xcc@{}}
\toprule
Method (backbone) & success & budget\\
\midrule
\multicolumn{3}{@{}l}{\emph{ALFWorld, token plane (Qwen2.5-1.5B / 7B)}}\\
PPO & $54.4$ / $80.4$ & 150 it\\
RLOO & $69.7$ / $75.5$ & 150 it\\
GRPO & $72.8$ / $77.6$ & 150 it\\
GiGPO & $86.7$ / $90.8$ & 150 it, 2 / 4 GPU\\
GiGPO & $86.9$ / $98.1$ & 500 it\\
DARS (token SOTA) & $96.9$ / $98.6$ & 500 it\\
\midrule[0.2pt]
\multicolumn{3}{@{}l}{\emph{ALFWorld, decision plane (Qwen2.5-1.5B / 7B)}}\\
\ldm{} & 82.5 / 90.3 & 150 / 100 it, 11--17 h$^{\S}$\\
\textbf{\ldm{} distill+RL} & \textbf{100} / \textbf{100} & 100 it, 2.1--5.1 h$^{\P}$\\
\midrule
\multicolumn{3}{@{}l}{\emph{Sokoban (Qwen2.5)}}\\
RAGEN, 0.5B (1 act./turn) & 12.1 & $\le$200 it\\
ActFocus (PPO), 0.5B & 34.0 & 200 it, 4 GPU\\
\ldm{}, 0.5B & $74.9\pm11.5$ & 1000 it, 6.0--6.3 h$^{\S}$\\
\textbf{\ldm{}, 1.5B} & $\mathbf{77.2\pm5.2}$ & 1000 it, 3.5--4 h\\
\ldm{} distill+RL, 0.5B & $75.7\pm4.0$ & 150 it, 0.6--2.5 h$^{\S\ddagger}$\\
\ldm{} distill+RL, 1.5B & $75.5\pm3.4$ & 150 it, 0.8--2.5 h$^{\S\ddagger}$\\
\midrule
\multicolumn{3}{@{}l}{\emph{Sokoban, visual (Qwen2.5-VL-3B, verl-agent, image only)}}\\
GRPO & 67.1 & 200 epochs, 32 GPU\\
GiGPO & 81.0 & 200 epochs, 32 GPU\\
SPEAR (Dr.BoT, GiGPO) & $\VlPublishedSpearFinal$ & 200 epochs, 32 GPU\\
\ldm{} & $\VlPureTableFinal$ & \VlPureBudget$^{\S}$\\
\VlSpearRows
\bottomrule
\end{tabularx}
\par\smallskip\begin{minipage}{\columnwidth}\scriptsize
Sources: \citet{feng2025gigpo,chen2026dars,he2026actfocus,spear}; ActFocus is PPO-based; interfaces and visual recipe details in Appendix~\ref{app:visual}. LDM text Sokoban: final checkpoints, mean $\pm$ s.d.\ over three runs (Section~\ref{sec:distill}). LDM visual Sokoban: 128 fixed puzzles, $T{=}0.4$, $\le$15 actions; means average decoding streams within training seeds. \VlSpearMainSelectionNote{} Visual rows denote bundled recipes; GPU-h are elapsed times on shared hosts and charge prior search. $^{\S}$Shared node. $^{\P}$Student RL after BC from Qwen3.5-2B (teacher RL 8.0 h). $^{\ddagger}$Same, Qwen3.5-4B (2.9 h).
Exact overlap with the allowed training pool: 107/128 benchmark puzzles (Appendix~\ref{app:visual}).
\end{minipage}
\end{table}

The retrospective maximum on visual Sokoban is $\VlSpearTopResult\%$ (\VlSpearTopMethod{}); SPEAR reports 87.9\% under the 128-puzzle protocol. \VlSpearEvidenceNote{}

%% file: sections/credit.tex
\section{An Exact Credit Audit}
\label{sec:credit}

How accurate is the credit that a decision-plane update applies, and how does the token plane compare? On FrozenLake (4$\times$4, slippery, 10 steps) dynamic programming makes the decision-plane audit exact; token-plane quantities are estimated.

\paragraph{Protocol.}
Prompts are history-free (map, cell, menu), so the policy is Markov in (map, cell). For 16 held-out maps, dynamic programming over all cells and remaining horizons $h\le10$ gives the exact advantage $A^\pi_h(s,a)$ and objective $J$ (mean success probability), and, for the \ldm{}, the exact policy gradient $g^\star$ in one backward pass. From independent batches of $G{=}8$ episodes per map (about 800 decisions) we form sampled gradients with GRPO, GiGPO and exact (``oracle'') advantages. We report their cosine to the reference gradient, sign agreement with exact advantages, and the improvement $\Delta J$ after one SGD step scaled to a mean decision-distribution KL of 0.01. In the token plane the semantic policy is estimated from 32 completions per cell. Appendix~\ref{app:credit} gives all details.

\begin{table}[t]
\caption{Credit audit on FrozenLake (Qwen2.5-0.5B at the start of RL, 16 maps; batches of 128 episodes: decision plane \AuditNLdm, token answer-only \AuditNAns, token rationale \AuditNCot{} from shards; mean $\pm$ s.e.\ over batches). $\cos$: cosine to the reference gradient (exact for the decision plane; a Monte Carlo plug-in for the token plane, so token cosines are not comparable across planes). Sign: agreement of estimated with exact advantage (token plane: plug-in oracle from the estimated semantic policy) where $|A|>0.02$. $\Delta J$: improvement after one step at mean KL 0.01 (exact for the decision plane, re-estimated for the token plane). Dec.\ share: fraction of an update's absolute log-probability change at the answer-label token. Token rows: whole-string / token-mean score. Details in Appendix~\ref{app:credit}.}
\label{tab:credit}
\centering\footnotesize
\setlength{\tabcolsep}{2.5pt}
\begin{tabular}{@{}lcccc@{}}
\toprule
advantage & $\cos$ & sign & $\Delta J\,(\times10^{-3})$ & dec.\ share\\
\midrule
\multicolumn{5}{@{}l}{\emph{Decision plane} ($J{=}\CreditLdmJ$)}\\
\quad oracle & \CreditLdmOracleCos & 1.00 & \CreditLdmOracleDJ & 1.000\\
\quad GRPO & \CreditLdmGrpoCos & \CreditLdmGrpoSign & \CreditLdmGrpoDJ & 1.000\\
\quad GiGPO & \CreditLdmGigpoCos & \CreditLdmGigpoSign & \CreditLdmGigpoDJ & 1.000\\
\midrule
\CreditTokRows
\bottomrule
\end{tabular}
\end{table}

\paragraph{Where credit lands.}
With exact advantages, a single batch of about 800 sampled decisions gives a decision-plane gradient whose cosine with the exact policy gradient is \CreditLdmOracleCos. Every one of the 32 steps improves the exact objective, by \CreditLdmOracleDJ$\times10^{-3}$ on average at KL 0.01. With exact advantages, the aggregate update is thus well aligned and improving in every audited trial; the remaining misalignment reflects sampling noise over which decisions were visited. \CreditTokText

On one fixed enumerable text policy, marginalising its rationale aligns the gradient better (\EstCosWins), with no detected one-step gain and a small training gain (area \InvPooled, $t{=}\InvPooledT$) that a pre-registered replication did not confirm; after training, realisation noise persists in a Sokoban rationale policy (\RvSokTrCot\%) and is near zero for answer-only policies (Appendix~\ref{app:estimator}).

\paragraph{When credit is earned.}
The larger error source in the decision plane is advantage estimation. Estimated advantages (group-centred, so signs may legitimately differ) agree in sign with exact advantages on only \CreditLdmGrpoSign{} of the decisions where $|A^\pi_h|>0.02$ (about 42\% of all decisions). About 46\% of decisions get no credit because whole groups succeed or fail together. With these estimates the cosine drops to \CreditLdmGrpoCos--\CreditLdmGigpoCos, yet the updates still recover \CreditRecover{} (GRPO) and \CreditRecoverG{} (GiGPO) of the oracle improvement per step. Credit \emph{placement} thus carries no realisation noise in the decision plane, while \emph{estimation} (which decisions deserve credit) accounts for most of the remaining gap to the oracle.

%% file: sections/distill.tex
\section{Distillation across Models and Planes}
\label{sec:distill}

\subsection{Across model families}
Qwen2.5-1.5B needs 11--14 GPU-h to reach 80--86\% on ALFWorld with its own RL. Policy distillation followed by RL \citep{rusu2016policy,ross2011dagger} transfers choices on a shared menu across families and tokenizers. We roll out a Qwen3.5 teacher \ldm{} on training tasks, keep decisions from successful episodes, re-render freshly shuffled menus, and train a Qwen2.5 student with two epochs of cross-entropy before decision-level RL.

\begin{table}[t]
\caption{Distill + RL. Students are Qwen2.5, teachers are Qwen3.5 \ldm{}s trained with decision-level RL. Test success (ALFWorld: 134 unseen, greedy; Sokoban: 512 seeds, greedy; Countdown-after: 1,024 items, greedy). GPU-h counts student RL only; teacher RL cost in brackets.}
\label{tab:distill}
\centering\small
\setlength{\tabcolsep}{3pt}
\resizebox{\linewidth}{!}{\begin{tabular}{@{}lcccc@{}}
\toprule
Task (student) & teacher & pure RL & distill & distill+RL\\
\midrule
ALFWorld (1.5B) & 98.5 [8.0] & 79.9 / 85.1 & \BcAlf{} / 99.3 & \textbf{100 / 100}\\
 \ \ GPU-h & & 11.2 / 14.3 & -- & 2.1 / 2.6\\
ALFWorld (7B) & 98.5 [8.0] & 90.3$^{100}$ & 100 & \textbf{100}\\
 \ \ GPU-h & & 17$^{\S}$ & -- & 5.1\\
Sokoban (1.5B) & 74.2 [2.9] & $77.2\pm5.2^{1000}$ & \BcSokQfifteen & $75.5\pm3.4$\\
 \ \ GPU-h & & 3.5--4$^{e}$ & -- & 0.8--2.5$^{\S}$\\
Sokoban (0.5B) & 74.2 [2.9] & $74.9\pm11.5^{1000}$ & \BcSokQfive & $75.7\pm4.0$\\
 \ \ GPU-h & & 6.0--6.3$^{\S}$ & -- & 0.6--2.5$^{\S}$\\
Countdown-after (1.5B) & 67.4 [5.8] & 57.9 & 69.3 & \textbf{70.2}\\
 \ \ GPU-h & & 2.7 & -- & 2.6\\
\bottomrule
\multicolumn{5}{p{0.95\columnwidth}}{\footnotesize $^{\dagger}$A100. Countdown-after: 400 RL iterations, same recipe; BC on 24k teacher episodes. Sokoban: final checkpoints, mean $\pm$ s.d.\ over training runs (BC-only column: one run); distill+RL uses 150 student RL iterations from one teacher corpus (BC and RL seeds vary). $^{1000}$1,000 RL iterations, learning rate $10^{-4}$ (for 1.5B chosen on validation over $\{3\times10^{-5},10^{-4}\}$); validation-selected checkpoints give $81.8\pm2.8$ (1.5B) and $78.6\pm6.2$ (0.5B). $^{e}$Exclusive H100, estimated from late-training iteration time; logged times of these co-located runs are higher. $^{100}$Greedy, after 100 iterations. $^{\S}$Shared node.}
\end{tabular}}
\end{table}

Table~\ref{tab:distill} shows where demonstrations help. On ALFWorld BC alone gives 99.3--100\% on unseen games and RL retains it at 2.1--2.6 GPU-h, far less than the student's own RL needs. On Sokoban, 150 student RL iterations bring both students to about 76\%, a level pure RL needs about 1,000 iterations to reach. On Countdown-after, filtering successful teacher episodes lets BC exceed its teacher (69.3 vs.\ 67.4\%), and RL preserves it (70.2\%). Appendix~\ref{app:distill} gives the data budgets and transfer details.

\subsection{Back to the generative LLM}
\label{sec:l2l}
\rlid{} also post-trains ordinary generative LLMs. The 100\% ALFWorld 1.5B \ldm{} of Table~\ref{tab:distill} plays one episode per training game, and sequence-level SFT teaches a generative LLM to write the chosen command as free text, \texttt{<action>command</action>}, parsed by verl-agent's projection (Appendix~\ref{app:l2l}).

\begin{table}[t]
\caption{\rlid{} $\rightarrow$ generative LLM: ALFWorld success (\%), Qwen2.5-1.5B student unless named. Greedy, except the no-memory row at $T{=}0.4$. Seeds and sampled reads in Appendix~\ref{app:l2l}.}
\label{tab:l2l}
\centering\small
\setlength{\tabcolsep}{4pt}
\begin{tabular}{@{}lcc@{}}
\toprule
Student & unseen & seen\\
\midrule
Memory prompt & \textbf{100} & \textbf{100}\\
Memory prompt, no action list & 99.3 & \textbf{100}\\
No memory & 90.5 & 97.1\\
Granite-3.3-2B, memory prompt & \textbf{100} & \textbf{100}\\
SmolLM2-1.7B, memory prompt & 96.3 & 99.3\\
Qwen2.5-0.5B, memory prompt & 98.5 & \textbf{100}\\
\midrule
GiGPO (published) & 86.7 & --\\
Token RL, same LLM and runner & 86.6 & --\\
\bottomrule
\end{tabular}
\end{table}

With compiled memory the generative student solves every unseen and seen game, both greedily and at $T{=}0.4$. It keeps 99.3\% unseen and 100\% seen success without the admissible-action list, so deployment needs no menu. Without memory it reaches 90.5\% at $T{=}0.4$, against GiGPO's published 86.7\%. The transfer holds across model families and sizes: Granite-3.3-2B reaches 100\%, SmolLM2-1.7B 96.3\% and Qwen2.5-0.5B 98.5\% unseen (Table~\ref{tab:l2l}).

The route is also cheaper than token-level RL. Including teacher training, the generative student reaches 99.3\% unseen success in 12.7 H100 GPU-h, versus 86.6\% in 17.3 GPU-h for token-level RL of the same LLM with the same prompt and runner. Given a trained \ldm{}, collection and 100 SFT steps (4.7 minutes) cost 0.47 GPU-h and give 100\% unseen success. Distilling directly from the Qwen3.5-2B \ldm{} also gives 99.3\%, and GSM8K and ARC-C scores are preserved (Appendix~\ref{app:l2l}).

%% file: sections/related.tex
\section{Related Work}
\label{sec:related}

\paragraph{Language models as policies over admissible actions.}
SayCan \citep{ahn2022saycan} scores actions without RL on the language model. GLAM \citep{carta2023glam} and TWOSOME \citep{tan2024twosome} fine-tune language models with online RL over admissible actions., scoring each action string by token likelihood; we read one categorical from a single prefill. Verbalised action masking \citep{vam2026} and POAD \citep{wen2024poad} keep generation and token credit.
 Large decision models \citep{chen2026ldm} and other supervised single-token readouts \citep{jev,nimble,koa2026,openjev}, constrained decoding and policy distillation also exist; \rlid{} adds the credit identity, controlled RL comparisons, an exact audit and distillation in both directions.

\paragraph{RL for language agents and credit.}
Group-baseline token RL \citep{shao2024deepseekmath} extends to agents in RAGEN \citep{wang2025ragen}, ArCHer \citep{zhou2024archer} and GiGPO \citep{feng2025gigpo}, whose decision-level anchor advantages we reuse; DARS \citep{chen2026dars} shapes step credit by task dependencies. Proposition~\ref{prop:rb} is a Rao--Blackwell argument for score-function estimators \citep{williams1992reinforce,sutton2000pg}.

%% file: sections/conclusion.tex
\section{Conclusion}
\label{sec:conclusion}
\rlid{} gives RL one prefill per decision, no realisation noise and no invalid outputs. Its policies distil back into ordinary generative LLMs that act without an admissible menu, hold their success under sampling and are reached at lower measured cost than token-level RL, so \rlid{} serves the post-training of both \ldm{}s and LLMs.

%% file: sections/appendix.tex
\section{Proof of Proposition~\ref{prop:rb}}
\label{app:proof}
Assume the parser $g$ does not depend on $\theta$, $p_\theta(y\mid x)$ is differentiable in $\theta$ with finite support or dominated derivatives, and the estimators have finite second moments. Write $p_\theta(y\mid x)=\pi_\theta(a\mid x)\,p_\theta(y\mid x,a)$ with $a=g(y)$ and $p_\theta(y\mid x,a)=p_\theta(y\mid x)/\pi_\theta(a\mid x)$ on $\{y:g(y)=a\}$. Taking logarithms and gradients gives the decomposition. For fixed $(x,a)$ with $\pi_\theta(a\mid x)>0$, $\sum_{y:g(y)=a}p_\theta(y\mid x,a)=1$ for all $\theta$, so $\E[\nabla\log p_\theta(y\mid x,a)\mid x,a]=\sum_y\nabla p_\theta(y\mid x,a)=\nabla 1=0$. Let $s_d=\nabla\log\pi_\theta(a\mid x)$ and $s_r=\nabla\log p_\theta(y\mid x,a)$. If $\hat A$ is conditionally independent of $y$ given $(x,a)$, then $\E[\hat A(s_d+s_r)\mid x,a,\hat A]=\hat A s_d$, so both estimators have the same mean. The difference of second moments is $\E[\hat A^2(2s_d^\top s_r+\|s_r\|^2)]$; the cross term vanishes by conditioning on $(x,a,\hat A)$. Hence $\operatorname{tr}\operatorname{Cov}$ differs by $\E[\hat A^2\|s_r\|^2]$.

\emph{When does the independence hold?} In our runner the history shown at later steps contains the executed action string and the observation, never the generated rationale. Future states, rewards and other episodes therefore depend on $y$ only through $a$, and so does any advantage computed from them, including GRPO, GiGPO and oracle advantages. If rationales were fed back into later prompts, future states would depend on $y$ beyond $a$, the conditional-independence assumption would fail, and the proposition would not apply without further assumptions.

\emph{Invalid outputs and format penalties.} In our token-plane runner an unparseable string executes a uniformly random action and receives a penalty. Rewards then depend on $y$ through the \emph{outcome class} $c=g'(y)\in\{1,\dots,k\}\cup\{\bot\}$ (a parsed label, or invalid) together with an execution kernel that maps $\bot$ to a random action. The proposition holds verbatim with $a$ replaced by $c$: $p(y\mid x)=\pi(c\mid x)p(y\mid x,c)$, the realisation score $\nabla\log p(y\mid x,c)$ has zero conditional mean, and any advantage that depends on $y$ only through $c$ and the execution randomness is conditionally independent of the realisation given $c$. The decision plane has no $\bot$ class.

\paragraph{Interpretation of the identity.}
Proposition~\ref{prop:rb} is an identity \emph{within one policy}: for a fixed text policy, its own score-function gradient equals the gradient of its induced decision policy plus a zero-mean term. That term assigns the decision's advantage to \emph{how} the action was verbalised: rationale tokens, formatting, and which of several strings parse to the same action. Estimating $\nabla\log\pi_\theta(a\mid x)$ instead removes this output-realisation sampling noise without changing the expected update. A free-form token policy cannot do this, because its marginal over strings is intractable. An \ldm{} (or, equivalently, generation constrained to a single legal label) parameterises $\pi_\theta(a\mid x)$ directly, so the realisation term is identically zero. This does not mean that an advantage moves only the evaluated decision's probability: parameters are shared across states, as in any policy gradient. The proposition does \emph{not} compare an \ldm{} with a token policy of different parameters: the two planes induce different policy classes (rationales add computation), and our token-plane baseline uses token-averaged PPO rather than the whole-string score. These are empirical questions, which Sections~\ref{sec:results} and \ref{sec:credit} address.

\section{Memory and History on ALFWorld}
\label{sec:analysis}
A single-pass policy sees only what the prompt contains. Appendix Table~\ref{tab:memory} crosses history length with the compiled memory on ALFWorld (Qwen3.5-2B, same optimiser settings, 150 iterations); dumped trajectories suggest why it helps. Memory-free 2-step failures in the inspected dumps are mostly 50-step exploration loops: the agent cycles through three or four receptacles with high confidence (repeating states for 31--36 of 50 steps) while unvisited cabinets stay closed. The compiled memory lists visited and unvisited receptacles and makes such loops visible in one line. In a matched comparison (same recipe and compiler, 2-step history) the memory raises success from 52.2\% to 79.9\% after 40 iterations (different evaluation menu orders) and from 91.8\% to 97.8\% at the end. State tracking moves into the compiler; the policy keeps only the choice.

\begin{table}[H]
\caption{ALFWorld (Qwen3.5-2B, 134 unseen games): history length $\times$ compiled memory. Final test success, sampled / greedy; in parentheses, sampled success at iteration 40. The no-memory runs predate several compiler changes (history as (action, resulting observation) instead of (observation, action), 52 instead of 62 labels, different menu truncation and anchor keys), so this table compares compiler and recipe versions rather than isolating memory; \NoMemText}
\label{tab:memory}
\centering\small
\begin{tabular}{@{}lcc@{}}
\toprule
 & no memory & compiled memory\\
\midrule
2-step history & 77.6 / 70.9 (45.5) & \textbf{97.8 / 98.5} (\textbf{79.9})\\
6-step history & 93.3 / 92.5 (45.5) & \AlfSixShuf{} (44.0)\\
\bottomrule
\end{tabular}
\end{table}

\section{Environments}
\begin{table}[H]
\caption{Environments. $k$ is the menu size; ``steps'' is the per-episode decision budget.}
\label{tab:envs}
\centering\small
\begin{tabular}{@{}lccl@{}}
\toprule
Env & steps & $k$ & menu source\\
\midrule
ARC-C & 1 & 3--5 & answer options\\
Countdown & 3 & $\le$36 & LLM-written compiler\\
Sokoban & 10 & 4 & native moves\\
FrozenLake & 10 & 4 & native moves\\
ALFWorld & 50 & 5--62$^{\ddagger}$ & admissible commands\\
\bottomrule
\end{tabular}

\smallskip\par\begin{minipage}{0.8\linewidth}\raggedright\footnotesize $^{\ddagger}$The retained menu holds at most 62 single-token labels. When ALFWorld's admissible commands exceed 62, generic commands are dropped first (\texttt{examine}, then \texttt{look}, \texttt{inventory}, \texttt{close}), never by reward (\texttt{ldmrl/core.py}); in the reported ALFWorld runs truncation affects 0.2--0.7\% of all decisions over a whole run (training and evaluation, cumulative \texttt{trunc\_rate}); per-iteration rates are higher in some iterations and are not separately logged. If the priority removals do not suffice, the remaining menu is cut positionally (\texttt{keep[:62]}).\end{minipage}
\end{table}

\section{Stability Lessons and Ablations}
\label{app:lessons}
\LessonsTable

\section{Qualitative Decision Statistics}
\label{app:qual}

In dumped episodes on unseen ALFWorld games (memory-free policies), the base Qwen3.5-2B \ldm{} solves 1 of 30 episodes, using 48.6 of 50 steps on average. Its mean chosen-action probability is 0.33, and 79\% of its decisions have probability below 0.5. After RL the mean chosen probability is 0.75--0.87, only 8--22\% of decisions fall below 0.5, and episodes take 10.8--16.1 steps at 82--93\% success. The decision entropy of the 1.5B student falls from 1.04 to 0.27 nats under pure RL, and to 0.07 nats under distill + RL (Section~\ref{sec:distill}).

\section{Failure Modes by Plane}
\paragraph{Failure modes that only exist in the token plane.}
\TokFailText

\paragraph{Label bias (both planes).}
Label bias affects RL over labelled menus in both planes (Section~\ref{sec:results}). Pretrained backbones put content-independent mass on early letters, and RL with unshuffled menus amplifies it to near-deterministic collapse onto one label. Shuffling menus during training makes a content-driven policy's label marginal uniform. The marginal regulariser or the calibration in Eq.~\eqref{eq:ldm} then removes the residual preference. On fixed four-action menus (Sokoban) the regulariser gives 38--55\% with Qwen2.5-0.5B, against 17\% with calibration. On variable-size menus (ALFWorld) we use calibration instead (Appendix~\ref{app:lessons}). Appendix~\ref{app:lessons} gives the remaining ablations.

\section{Run Index and Hyperparameters}
\label{app:hparams}
Most table numbers are generated by \texttt{paper/extract\_finals.py} and \texttt{paper/make\_macros.py} from run logs and audit/benchmark artifacts; numbers in the narrative (and a few constants in \texttt{make\_macros.py}, e.g.\ the 0.12 GPU-h ARC cost) were checked against the same artifacts by hand. BC-only results come from the evaluation files of the distillation pipeline. Table~\ref{tab:runs} maps entries to runs. \emph{Token-plane sampling.} Every token-plane rollout samples at temperature 1 with top-$p{=}1$, no top-$k$ and no repetition penalty, so the sampled text comes from the distribution whose log-probabilities enter the PPO ratio. Qwen2.5's default generation configuration (repetition penalty 1.1, top-$k$ 20) would otherwise apply silently; earlier versions of the Sokoban token arms and of the Countdown-after token baseline (\texttt{tokrl\_cda\_q25\_1.5b}, 25.2\%) inherited it and are superseded by the reruns reported here (\texttt{mSf\_*}, \texttt{tokrl\_cdafix\_q25\_1.5b\_s\{17,1,2\}}). \emph{Recipe sensitivity (Countdown-after).} The first \ldm{} run predates two code changes to the label regulariser: a per-iteration (``global'') variant became the default, and zero-advantage rows were fed to the regulariser. Replicates with both changes (\texttt{--marg-mode global}, all rows; three seeds) over-regularised to near-uniform decisions and reached 7.4, 7.8 and 12.9\% at 12.8k episodes; a micro-batch variant on all rows showed the same entropy rise and was stopped after 12--13 iterations. The reported replicates restore the per-micro-batch penalty on non-zero-advantage rows (\texttt{--marg-mode micro --marg-rows adv}), which is our intended reconstruction of the original recipe; the original code revision was not retained. Metrics of the unsuccessful attempts are in \texttt{results/metrics\_failed}. The Countdown-after comparison is therefore between the tested recipes. \emph{Evaluation randomness.} An invalid token-plane output is replaced by a uniformly random menu item. In the reported runs this fallback drew from the training RNG during evaluation unless menus were shuffled, so a token-plane evaluation could shift later training randomness and greedy token evaluations are not fully deterministic (the decision plane has no invalid outputs and is unaffected). The released code gives evaluation its own fixed-seed Python RNG and restores the PyTorch RNG state after seeded sampling; apart from \texttt{mC\_rcd\_s1} (run after the Python-RNG fix), reported token-plane runs predate these fixes and were not rerun.
\RunIndexTable

\ValSelTable

\paragraph{Fixed label order on Sokoban.}
Sokoban's four native moves always appear in the same order, so their labels can stay fixed. On Qwen2.5-0.5B (300 iterations, same host), a fixed order reaches 70.1\% eval and 23.2\% OOD in 0.63 GPU-h, against 64.5\% and 20.7\% in 0.74 GPU-h with shuffled labels; at iteration 100 the fixed order leads 36.7\% to 21.5\%.

\paragraph{Visual Sokoban and published comparisons (Table~\ref{tab:published}).}
\label{app:visual}
The published comparisons are contextual, not matched protocols. ALFWorld decision-plane rows use 134 unseen games: Qwen2.5 greedy (1.5B / 7B), Qwen3.5 sampled / greedy. PPO--GiGPO at 150 iterations come from \citet{feng2025gigpo}; GiGPO and DARS at 500 from \citet{chen2026dars}, who use 128 games, one training run per arm and a mean over decoding draws (7B validation-selected).

\begin{table}[H]
\caption{Additional ALFWorld backbones: success (\%) on 134 unseen games, sampled / greedy, and per-run budget.}
\label{tab:alf-additional-backbones}
\centering\footnotesize
\begin{tabularx}{\textwidth}{@{}Xcc@{}}
\toprule
Method (backbone) & success & budget\\
\midrule
\ldm{}, Qwen3.5-2B & \AlfQtwoB & 150 it, 8--9 h\\
\ldm{}, Qwen3.5-4B & 99.3 / 99.3 & 150 it, 11 h\\
\bottomrule
\end{tabularx}
\end{table}

Visual Sokoban follows the verl-agent generator: $6\times6$, one box, search depth 30, native rewards and at most 15 primitive actions. The actor sees the rendered image and four directional actions, with no text grid or observation history. SPEAR describes RGB and textual inputs; its released RGB path pairs images with static instructions and action-format text, without a textual board rendering. Our language-model LoRA and one-pass menu-label readout differ from SPEAR's full-parameter token policy and reasoning interface. Benchmark seeds are the fixed first draw of 128 validation seeds, rather than redrawn at each validation. Final adapters of \texttt{vaS\_vl3b\_rl} and its seeds 1--2 train for 200 iterations of $32\times8$ episodes. Each training seed is evaluated at $T=0.4$ with three decoding streams (0, 1, 2); their within-seed means are 82.8125, 83.3333 and 88.2813\%, giving $\VlPureFinal\%$ across training seeds. The historic stream-0-only mean is $85.2\pm2.7\%$. The best observed seed is a retrospective maximum, not a recipe selected independently of the benchmark, and does not establish a reliable lead over SPEAR's reported 87.9\%. The text-teacher result below retains the original single-run, stream-0 final evaluation.

GiGPO's reported seed variation is over inference seeds. Our pure-RL LDM baseline varies three independent training seeds and averages three decoding streams within each seed; these two sources of variation are reported separately.

\paragraph{Recipes and attribution.}
The short Table~\ref{tab:published} labels denote bundled recipes. Continuations load adapter weights with a fresh AdamW optimizer, learning rate $5\times10^{-5}$ and cosine decay to 20\% of that rate. SPEAR replay is our decision-level adaptation: capacity 2,048, batch 512, maximum age 10 iterations, peak weight 0.1 and warmup 30 iterations (100 from base). Admission requires positive original GiGPO credit and episode return above the median of up to 10,240 puzzle-group mean returns; this history covers the whole 200-iteration run. Replay arms also widen the on-policy upper PPO clip to 0.28. Curriculum samples at most half the batch by success EMA $p$ (decay 0.9), with priority $\min(1.31,\,0.1+0.4(1-p)+4p(1-p))$, and also adds normalized-entropy floor 0.15 with coefficient 0.2. Curriculum from base adds the same learning-rate decay to the otherwise constant-$10^{-4}$ baseline; replay from base uses constant $10^{-4}$. The unnormalized-advantage arm also sets $\gamma=1$, learning rate $5\times10^{-5}$, cosine decay and upper clip 0.28. GiGPO's training-only anchor key is the text-grid state, equivalent to grouping identical rendered images; it is never passed to the actor.

Replay eligibility is sparse in the replay-only continuation (20 replay decisions at the final iteration). Its four extra AdamW steps per iteration reuse the on-policy optimizer: stored momentum can move weights even when replay gradients are tiny. We have no matched clipping/optimizer-step control, so gains do not isolate self-imitation. The continuation control is 7.0 points above its original seed-0 final; replay, curriculum and their combination add 2.1, 0.8 and 2.9 points over that control. These are exploratory point estimates, with no paired per-puzzle significance test or independent-seed evidence yet for the replay/combination recipes. Three decoding streams do not substitute for independent training seeds. Shared-host elapsed GPU-hours are not isolated throughput measurements.

The standard seed split does not guarantee distinct initial boards: the 65,535 training seeds generate 21,495 distinct boards, and 104 of the 124 distinct benchmark boards occur in that pool. After excluding the 256 selection seeds, 103 benchmark boards still occur in the remaining training pool (107 of the 128 benchmark seed draws). Those selection seeds produce 249 distinct boards, of which 205 occur in the remaining training pool and six occur in the benchmark. Selection is therefore held out by seed, not by board. These counts compare allowed training pools, not the episodes actually sampled by each run.

The separately frozen 1,024-puzzle holdout excludes exact initial-state matches (walls, targets, boxes and player) with training, benchmark and selection. It retains symmetry overlap: 728 of 1,024 puzzles (679 of 971 classes) are rotated or reflected copies of boards in the training pool. Also ignoring player position within the same reachable component raises overlap to 914 of 1,024 puzzles (699 of 807 classes), though this can change action costs under the 15-action budget. For the benchmark, 122 of 128 seed draws match the training pool up to symmetry, and 127 of 128 under the component comparison. Neither split establishes generalization to novel puzzle structures. The audit covers initial states; intermediate-state overlap and pretrained-model contamination are unmeasured. The seed-list SHA1 remains \texttt{b8218d00b281421a16ea6cb0547053a42cebf4c6}; the read-only audit is \texttt{results/sokoban\_visual\_spear/equivalence\_board\_audit.json}.

On the holdout, at $T{=}0.4$ with validation-selected checkpoints (mean $\pm$ s.d.\ over three training runs; benchmark in parentheses), teacher-free RL reaches $76.7\pm1.6$ ($83.7\pm1.8$), text-teacher BC alone $75.2\pm0.3$ ($83.1\pm0.7$), and text-teacher BC then RL $81.2\pm1.6$ ($85.9\pm1.6$): every model loses 4.7--7.9 points, and the text teacher's benefit is larger there (+4.5 vs.\ +2.2). The holdout filter rejected 4,828 of 5,881 scanned candidates for training-board matches, shifting the sample toward rarer generator outputs; the score drop does not isolate memorization. New replay/curriculum experiments log only selection-puzzle evaluations during training and evaluate the benchmark after training completes. The current holdout campaign evaluates frozen final adapters without using holdout outcomes for optimization or recipe/checkpoint selection. Replay from base was not declared in that five-variant holdout campaign. Replay uses the actor's own decisions and episode outcomes; it supplies no solver or text-teacher labels. Table~\ref{tab:published}'s generated rows require completed training, final evaluations, matching arguments and recorded initialization hashes. \VlSpearNote{}

\begin{table*}[t]
\caption{All completed visual Sokoban experiments, final checkpoints. Selection is greedy success (\%) on 256 reserved training seeds; benchmark success (\%) averages three decoding streams at $T=0.4$ within each training seed. Pilot recipes use one training seed; replicated results use three independent prefixes. Budgets charge the prior search and the new training.}
\label{tab:visual-pilots}
\centering\footnotesize
\begin{tabularx}{\textwidth}{@{}Xccc@{}}
\toprule
Method & selection & benchmark & budget\\
\midrule
\VlSpearAllRows
\midrule
LDM, retrospective best pure-RL seed & --- & $\VlPurePeak$ & \VlPurePeakBudget\\
LDM, privileged text-teacher BC+RL & --- & $\VlTeacherFinal$ & \VlTeacherBudget$^{*}$\\
\bottomrule
\end{tabularx}
\par\smallskip\begin{minipage}{\textwidth}\footnotesize
$^{*}$The teacher supplies text-grid decisions during BC; student inference is image only. Teacher rollouts and BC add 0.4 h to the student RL budget shown. The seed maximum is retrospective; the teacher result uses its original decoding stream. These two results are separate from recipe selection.
\end{minipage}
\end{table*}

\paragraph{Final checkpoints on the sealed holdout.}
Table~\ref{tab:visual-holdout-final} evaluates frozen final adapters on the separate 1,024-board holdout at $T=0.4$, with the same image-only interface and 15-action budget. New variants check checkpoint and source hashes before and after evaluation; complete records match the sealed seed-list hash and original training specification. Legacy pure-RL records verify the full protocol and checkpoint paths, but did not record checkpoint content hashes at evaluation. \VlHoldoutNote{}

\begin{table}[H]
\caption{Visual Sokoban on the 1,024-puzzle holdout, final checkpoints; success (\%). Exact initial-board duplicates are excluded; rotated and reflected copies remain. Baseline: mean $\pm$ s.d.\ over three training seeds. New variants: seed 0. Each seed averages three decoding streams. Only complete evaluations are shown.}
\label{tab:visual-holdout-final}
\centering\footnotesize
\begin{tabularx}{\textwidth}{@{}Xc@{}}
\toprule
Method & holdout\\
\midrule
\VlHoldoutRows
\bottomrule
\end{tabularx}
\end{table}

\section{Distillation Ablation}
\label{app:distill}
Table~\ref{tab:distillabl} compares hard and soft teacher labels, with and without an entropy floor during student RL, for the Countdown-after student of Section~\ref{sec:distill}, distilled from the first, small dump (3,000 teacher episodes) with an earlier RL recipe (learning rate $2\times10^{-5}$, calibration 0.5, no label regulariser, 150 iterations). With the current recipe for 400 iterations, RL from this BC adapter ends at 60.9\% and pure RL at 57.9\%. The main-text result uses a dump eight times larger.
\DistillAblTable

\paragraph{Across-family transfer details.}
\emph{ALFWorld.} The student's own RL stays below the teacher (79.9--85.8 vs.\ 98.5). Decision-level BC alone gives 99.3--100\% on unseen games for both Qwen2.5-1.5B and Qwen2.5-7B (from 15,224 teacher decisions), RL mostly retains it, and the student's RL budget falls from 11--14 to 2.1--2.6 GPU-h.
\emph{Sokoban: speed, not ceiling.} BC on 8,019 decisions of the Qwen3.5-4B teacher gives 70.3--74.0\%. Three distill+RL runs per student (one teacher corpus; BC and RL seeds vary; 150 iterations) end at $75.7\pm4.0$ (0.5B; runs 77.5, 71.1, 78.3) and $75.5\pm3.4$ (1.5B), with $21.4\pm4.2$ and $20.7\pm2.4$ on 8$\times$8 two-box maps (validation-selected). Pure RL needs about 1,000 iterations to get there but then matches or exceeds it: final $74.9\pm11.5$ (0.5B) and $77.2\pm5.2$ (1.5B), validation-selected $78.6\pm6.2$ and $81.8\pm2.8$, OOD $29.4\pm5.8$ and $28.8\pm5.3$. For 1.5B the learning rate was chosen before inspecting test results by mean validation success at the selected checkpoints ($10^{-4}$: 80.7; $3\times10^{-5}$: 70.2, test $73.3\pm11.8$ final). An earlier 1.5B pure-RL run at $10^{-5}$ for 300 iterations (19.7\%) was untuned and is superseded.
\emph{When pure RL is already strong.} On Countdown-after the student's own RL reaches 57.9\% after 400 iterations. With 3,000 teacher episodes BC gave \BcCd\% (Appendix Table~\ref{tab:distillabl}); with 24k sampled at temperature 1 (38.5k decisions from the 69\% that succeed), BC alone reaches 69.3\%, above the teacher's greedy 67.4\%: keeping only successful samples filters out the teacher's mistakes. RL then preserves it (70.2\%; on untouched problems 70.1 vs.\ teacher 65.9 and pure RL 56.6).

\section{Controlled Throughput}
Table~\ref{tab:throughput} gives the microbenchmark behind the cost-per-decision comparison of Section~\ref{sec:results}.
\begin{table}[H]
\caption{Controlled throughput on 256 Sokoban decisions (one H100 shared with other jobs; all policies timed back to back with identical micro-batch sizes; median of five repetitions, all repetitions are in the released artifact). Gen.: mean generated tokens per decision. Roll / train: seconds for rollout (generation and log-probabilities, no gradient) and for a training forward/backward pass. Dec/s: decisions per second over both phases.}
\label{tab:throughput}
\centering\small
\begin{tabular}{@{}llccccc@{}}
\toprule
Backbone & policy & gen. & roll (s) & train (s) & dec/s & \ldm{} speed-up\\
\midrule
\ThroughputRows
\bottomrule
\end{tabular}
\end{table}

\section{Credit Audit Details}
\label{app:credit}
\CreditDetails

\section{Estimator Test and Realisation Noise After Training}
\label{app:estimator}
\paragraph{Same policy, two estimators.} To separate the estimator from the policy class, we audit a fully enumerable text policy on FrozenLake (Qwen2.5-0.5B, LoRA rank 16; \texttt{ldmrl/estimator\_audit.py}, \texttt{results/est2}). At each state it first emits one of $R$ single-token rationales $r$, then an action label $a$: $p(r,a\mid x)=p(r\mid x)\,p(a\mid x,r)$, with no invalid strings. Its action policy is the exact marginal $\pi(a\mid x)=\sum_r p(r\mid x)p(a\mid x,r)$. On the same sampled $(s,r,a)$ and exact advantages we compare the whole-string estimator $W=\overline{A\,\nabla\log p(r,a\mid x)}$ with the marginalised estimator $M=\overline{A\,\nabla\log\pi(a\mid x)}$, which is what the decision plane computes for this policy. Both average over the random number of visited decisions, so they are ratio estimators with the same expectation but are not unbiased for $\nabla J$; their episode-normalised versions (sums divided by the number of episodes) are, and the error column uses those. Cosines and update directions are unaffected by this positive rescaling. Each update is line-searched to a semantic KL of 0.01 with a 3\% relative tolerance (up to 40 trials); achieved KL lies in \EstKLRange, and \EstMatchedRange{} of 16 batch pairs per setting have both updates within tolerance. The return comparison uses only those matched pairs; the exact change in return is computed by dynamic programming (16 maps, 16 batches of 8 episodes per map, parameters restored after each update).

\begin{table}[H]
\caption{Estimator test (entries W / M). Cosine and episode-normalised relative squared error to the exact gradient (all 16 batches); $\Delta J$: paired difference M$-$W of the exact return change ($\times10^{-3}$, mean $\pm$ s.e.) over batch pairs whose two updates both meet the KL tolerance, and the number of those pairs where M is better.}
\label{tab:estimator}
\centering\small
\begin{tabular}{@{}lcccc@{}}
\toprule
Policy & cosine & rel.\ error & $\Delta J$ (M$-$W) & M better\\
\midrule
\EstRows
\bottomrule
\end{tabular}
\end{table}

Marginalising over the rationale gives a gradient closer to the exact one (cosine higher in \EstCosWins{} settings, error lower in \EstErrWins). Its effect on the one-step return is not detectable: the paired differences are positive in \EstPositive{} of \EstN{} settings but within two standard errors in each, and pooled over all \EstPairs{} matched pairs the difference is \EstPooled$\times10^{-3}$ (\EstPooledPos{} positive; standard error clustered over the \EstClusters{} groups of pairs that share a batch random stream, since settings with the same seed reuse batch seeds). This is no detected gain, not evidence of equivalence. In this small, valid-output setting, removing realisation noise therefore yields a better-aligned gradient but no detected per-step return gain; the larger gaps of Section~\ref{sec:results} also involve invalid outputs, longer realisations and recipe effects.

\paragraph{Longer rationales and training.} We extend the same policy to rationales of $L$ tokens, each from the same $R{=}3$ choices, enumerating all $3^L$ paths exactly. Realisation noise grows with $L$ (one-step diagnostics with oracle advantages and episode-normalised errors; random init at scale 0.02, whereas training below uses 0.01, so these ratios do not measure noise along the trained GRPO trajectories): the ratio of the realisation term $\E\|g_W-g_M\|^2$ to the marginalised estimator's own error is \RatioDoneBase{} (base) and \RatioDoneInitone{} (random init) at $L{=}1$, against \RatioDthreeBase{} and \RatioDthreeInitone{} at $L{=}3$, and W's cosine to the exact gradient falls to \DepthDthreeInitoneCos{} (W / M). We then train W and M separately for 30 iterations from identical starting parameters (random LoRA init, scale 0.01) with identical per-iteration sampling seeds, keeping each KL-targeted step (semantic KL 0.01; the closest searched step is kept even when the 3\% tolerance is missed, so both arms take every step; mean achieved KL is balanced and the matched counts are in the table) and tracking exact $J$ by dynamic programming. Each seed is an independent pair of runs.

\begin{table}[H]
\caption{W vs.\ M over 30 KL-targeted training iterations (Qwen2.5-0.5B, FrozenLake, 16 maps, 8 episodes per map per iteration). Final $J$ (mean over seeds, W / M); paired differences M$-$W of final $J$ and of the area under the exact-$J$ curve, mean $\pm$ s.e.\ over seeds (positive seeds in brackets); matched steps out of all steps (W, M).}
\label{tab:train}
\centering\small
\begin{tabular}{@{}lccccc@{}}
\toprule
Advantages, rationale & seeds & final $J$ & $\Delta$ final $J$ & $\Delta$ area & matched steps\\
\midrule
\TrRows
\bottomrule
\end{tabular}
\end{table}

M has higher mean returns than W in every arm, with substantial seed uncertainty, and the effect is small: with GRPO-estimated advantages and three-token rationales the final-$J$ difference is \TrGrpoDthreeFinal{} (about \TrGrpoDthreeRel\% of W's final $J$; the area difference \TrGrpoDthreeAuc, positive in \TrGrpoDthreeAucPos{} seeds, has an interval that includes zero), with oracle advantages \TrOracleDthreeFinal{} and area \TrOracleDthreeAuc{} (\TrOracleDthreeAucPos), and the effect does not grow from one to three rationale tokens even though the realisation noise does. In this setting, removing realisation noise yields a consistently better-aligned gradient but only a small learning advantage that is not clearly distinguished from zero (seed-level 95\% $t$-intervals for the GRPO depth-3 differences: final $J$ \TrGrpoDthreeFinalCi, area \TrGrpoDthreeAucCi; oracle area \TrOracleDthreeAucCi). We therefore do not attribute the larger benchmark gaps to realisation noise alone.

\paragraph{Unparseable outputs.} Unparseable outputs, which execute a random action, are one failure mode of the token arms of Section~\ref{sec:results} (86\% of early Sokoban rationale decisions), although trained rationale policies become largely valid. We add this regime to the same within-policy test: the policy can also emit one of $N$ invalid label tokens (a fixed logit bias, learnable logits underneath, sets the initial invalid mass), and an invalid emission executes a uniformly random action, with no penalty. W scores the emitted string including an invalid label; M scores the exact probability of the executed action, $\pi(a\mid x)=\sum_{r}p(r\mid x)\,[p(a\mid x,r)+p(\mathrm{inv}\mid x,r)/4]$. Training as above (GRPO-estimated advantages, one-token rationales, 30 KL-targeted iterations, eight seeds per arm; endpoints fixed before the runs: paired differences in final $J$ and area under the $J$ curve).

\begin{table}[H]
\caption{W vs.\ M with unparseable outputs (GRPO advantages). Rows are named by the logit bias (low / high; nominal targets 30\% / 80\%); the actual mean initial invalid mass (\%) is given, and the number of invalid labels $N$ does not hold that mass fixed. Paired M$-$W differences in area under the exact-$J$ curve and in final $J$, mean with seed-level 95\% $t$-interval (positive seeds in brackets); mean achieved KL ($\times10^{-3}$), matched steps and skipped steps (non-finite or zero gradient), each W / M over 240 steps.}
\label{tab:invalid}
\centering\small\setlength{\tabcolsep}{4pt}
\begin{tabular}{@{}lccccccc@{}}
\toprule
Invalid labels & init.\ mass & seeds & $\Delta$ area & $\Delta$ final $J$ & KL & matched & skipped\\
\midrule
\InvRows
\bottomrule
\end{tabular}
\end{table}

M has the larger area in every arm. Pooled over all \InvPairs{} pairs, with uncertainty clustered by seed because arms share seeds, the area difference is \InvPooled{} ($t{=}\InvPooledT$, positive in \InvPooledSeeds{} seed clusters), about \InvPooledRel\% of W's area: removing realisation noise gives a statistically clear but modest learning advantage. The final-$J$ differences are not individually distinguished from zero, and the dose response is not significant (slope of the area difference in initial invalid mass \InvSlope, standard error clustered by seed). In one 4-token, high-bias seed W's policy collapsed onto invalid labels (95\% invalid mass), its gradient became non-finite and \InvFourEightyNonfinite{} of its updates were skipped, while M continued to improve; this seed is included as pre-declared, and excluding it leaves the pooled area difference at $+0.47\pm0.13$ (95\% interval $[+0.16, +0.77]$). In that 4-token, high-bias arm the line search was also unstable for W (mean achieved KL 0.0075 vs.\ 0.0101 for M, fewer matched steps, very large accepted step scales before the collapse), so its result is an advantage under this estimator and update procedure, not a clean equal-KL variance intervention. Without an invalidity penalty both estimators often drift towards high invalid mass, since an invalid output is then just a random action.

\paragraph{A pre-registered realisation-sampling intervention.} Finally we ran a design fixed in advance together with its success criterion (learning-curve area of M at least 10\% above W, a seed-level 95\% interval excluding zero, at least 7 of 8 seeds, and a realisation-averaged arm recovering at least half of the gap): three-token rationales, GRPO advantages, 20 KL-targeted updates, \WkN{} fresh seeds, three arms from identical starts: W; W$_4$, which keeps each decision's action and advantage but averages its whole-string score over four rationale realisations drawn from the exact posterior $p(r\mid x,a)$; and M. Mean achieved KL is 0.0100 in every arm with no skipped updates, but 11/160 (W), 14/160 (W$_4$) and 9/160 (M) updates miss the 3\% KL tolerance. M's area exceeds W's by \WkMWRel\% (absolute area difference \WkMWCi, \WkMWPos{} seeds), and W$_4$'s exceeds W's by \WkWfourWRel\% (\WkWfourWCi, \WkWfourWPos), so M lies below W$_4$; no final-$J$ difference was detected. The frozen shared-checkpoint diagnostics of the original design were not run. The criterion is not met: in this setting, reducing realisation noise does not produce a detectable learning benefit, and the paper's benchmark gains should not be attributed to it.

\paragraph{Realisation noise after training.} \texttt{ldmrl/realisation\_var2.py} (\texttt{results/rv2b}) repeats the measurement of Section~\ref{sec:credit} on 32 Sokoban and Countdown-after states per policy (32 scored completions each), for base and trained token policies (\texttt{mSf\_\{ans,cot\}\_s0}, \texttt{mC\_\{ans,cot\}\_s0}). States come from short random prefixes and are kept only if successor solvability differs across actions. The weights are an auxiliary solvability diagnostic, not the training advantages: $A(s,c)=v(s'_c)-\sum_{c'}P_{\mathrm{ref}}(c')v(s'_{c'})$, with $v$ the exact solvability of the successor within the remaining budget (Sokoban: BFS within the 10-step training budget; Countdown: exhaustive search) and $P_{\mathrm{ref}}$ estimated from 32 further, independent completions, so the weights do not depend on the scored completions. An invalid output executes a random action. Classes with one completion count as decision variance.

\begin{table}[H]
\caption{Realisation share (\%) of the score second moment for base and trained token policies. Raw: unweighted, all outputs; adv., valid: weighted by $A(s,c)^2$ and restricted to parseable outputs, with a 95\% bootstrap interval over states and the number of states that contribute non-negligibly; last column: token-mean score (as in token GRPO). Sequence-sum score otherwise.}
\label{tab:rv2}
\centering\small
\begin{tabular}{@{}lccccc@{}}
\toprule
Policy & valid (\%) & raw & adv., valid [95\% CI] & states & adv., valid, mean\\
\midrule
\RvRows
\bottomrule
\end{tabular}
\end{table}

Realisation noise survives training in a predominantly valid rationale policy on Sokoban (\RvSokTrCotValid\% valid; \RvSokTrCotCi\% of the advantage-weighted score). Trained answer-only policies produce predominantly canonical ``\texttt{Answer: X}'' outputs with approximately zero weighted within-class score variation on this panel; only \RvSokTrAnsContrib{} (Sokoban) and \RvCdaTrAnsContrib{} (Countdown) states contribute, bootstrap draws with no contributing state are discarded, and single-completion classes have no within-class term, so their 0\% intervals do not indicate population-level certainty. A trained Countdown rationale policy retains \RvCdaTrCotCi\% weighted but \RvCdaTrCotRaw\% unweighted: it is near-deterministic in its action on these states (the most frequent action takes 87\% of samples on average), so little weight falls on classes whose wording varies. Low weighted shares therefore do not imply that the text stopped varying, and the evidence is heterogeneous across tasks. The decision plane has no realisation term at any point of training.

\section{Frozen Countdown-after Protocol}
\label{app:protocol}
The protocol was written before any of its runs started (\texttt{paper/reviews/protocol\_r17.md}). Four arms share the runner, menus, rollout grouping, advantages, PPO, optimiser and 25.6k training episodes (100 iterations of $32\times8$): the \ldm{}; the hybrid (rationale of at most 64 tokens, then a constrained label with the \ldm{}'s regulariser; end-of-turn masked, and a malformed answer cue ends the rationale with its trigger token scored); answer-only and 30-word-rationale token policies. For each arm the learning rate was selected on the validation split (256 held-out training problems) from $\{2,5,10\}\times10^{-5}$ with seed 0 (ties to the smaller rate), giving $10^{-4}$ for the \ldm{}, hybrid and answer-only arms and $2\times10^{-5}$ for the rationale arm; seeds 1--3 were then run at the selected rate, and, in a second pre-registered stage, seeds 4--7 (eight seeds per arm). The same stage added a fifth arm, the \ldm{} without the label regulariser (learning rate selected the same way: $10^{-4}$), whose contrast with the unregularised token arms isolates the plane from the regulariser; one of its eight seeds (seed 6) failed with repeated out-of-memory errors on a shared GPU, leaving seven. Only final checkpoints are reported (evaluated greedily), with strict success (an unparseable output fails the episode). The untouched split holds 2,048 problems from the source pool that are in neither training nor test data (\texttt{scripts/make\_fresh\_cd.py}); it was first evaluated by these runs. By multiset of inputs and target, the original test split shares 32 of its 1,024 problems with training (a property of the original random split); the untouched and second sealed splits share none with training, test or each other (hashes and overlaps: \texttt{results/countdown\_split\_manifest.txt}).

\begin{table}[H]
\caption{Frozen protocol on Countdown-after (Qwen2.5-1.5B): strict success (\%), mean $\pm$ s.d.\ over eight seeds (seven for the unregularised \ldm{}); paired-by-seed differences with bootstrap 95\% intervals and paired-$t$ $p$-values; seconds per training iteration (rollout + update) with each run alone on one H100 (iterations 3--6 from the final checkpoint).}
\label{tab:protocol}
\centering\footnotesize
\setlength{\tabcolsep}{3pt}
\begin{tabular}{@{}lccccc@{}}
\toprule
Arm & test & untouched & contrast & test & untouched\\
\midrule
\ldm{} & $44.4\pm5.2$ & $42.2\pm5.3$ & & & \\
hybrid & $44.2\pm5.5$ & $41.9\pm4.8$ & \ldm{} $-$ hybrid & $+0.2\ [-4.5,+4.6]$, $p{=}0.93$ & $+0.3\ [-4.3,+4.4]$, $p{=}0.91$\\
answer only & $32.0\pm2.2$ & $30.3\pm2.2$ & \ldm{} $-$ answer & $+12.4\ [+9.4,+15.2]$, $p{<}0.001$ & $+11.8\ [+9.2,+14.2]$, $p{<}0.001$\\
30-word rationale & $30.7\pm2.5$ & $28.4\pm2.0$ & \ldm{} $-$ rationale & $+13.7\ [+9.1,+18.0]$, $p{<}0.001$ & $+13.7\ [+9.2,+17.9]$, $p{<}0.001$\\
\ldm{}, no label reg. & $33.1\pm2.0$ & $31.6\pm2.3$ & no reg.\ $-$ answer & $+1.1\ [-1.4,+3.4]$, $p{=}0.44$ & $+1.3\ [-1.4,+4.0]$, $p{=}0.41$\\
 & & & no reg.\ $-$ rationale & $+2.7\ [+0.3,+4.7]$, $p{=}0.07$ & $+3.6\ [+0.9,+5.8]$, $p{=}0.03$\\
 & & & \ldm{} $-$ no reg. & $+11.8\ [+7.4,+15.7]$, $p{=}0.002$ & $+10.9\ [+6.4,+14.6]$, $p{=}0.003$\\
\bottomrule
\end{tabular}
\end{table}
Per seed (seeds 0--7, test): \ldm{} 49.6, 47.9, 47.8, 37.9, 50.4, 43.3, 41.5, 37.3; hybrid 50.9, 46.7, 36.1, 39.9, 47.3, 42.8, 39.4, 50.7; answer only 33.8, 30.5, 34.6, 32.7, 34.1, 28.4, 32.2, 29.8; rationale 30.0, 27.1, 32.8, 32.7, 29.3, 27.7, 33.3, 33.0; \ldm{} without regulariser 35.4, 32.5, 29.9, 32.1, 33.3, 32.5, --, 35.7. Seconds per iteration: \ldm{} 25.6, hybrid 56.4, answer only 38.2, rationale 55.0.
Seed-0 runs at the selected rates (test / untouched): \ldm{} 49.6 / 47.2, hybrid 50.9 / 47.4, answer only 33.8 / 31.7, rationale 30.0 / 26.4 (strict). The hybrid generates 39--58 rationale tokens per decision after training; its rollout takes 41\,s per iteration against 11\,s for the \ldm{}, while updates cost the same (15\,s). Elapsed through the last training iteration on shared H100s, including periodic evaluation (excluding initialisation and final evaluation), the protocol runs took 1.65 (\ldm{}), 2.47 (answer only), 2.94 (rationale) and 3.33 (hybrid) GPU-h on average.

\paragraph{Plain Countdown.} Without afterstates the menu lists slot references (``combine slots 1 and 3 with $-$''), so a single pass must search without lookahead. Neither plane does well at 1.5B (Table~\ref{tab:plaincd}).
\begin{table}[H]
\caption{Plain Countdown (slot-reference menus): test success (\%), mean $\pm$ s.d.\ (seeds); episodes (k) and H100 GPU-h as decision / token. Same-runner rows use 512 test problems at 25.6k episodes. Full-expression GRPO (TRL) is a different interface, and its results use the first 512 test problems, the decision plane's all 1,024.}
\label{tab:plaincd}
\centering\footnotesize
\setlength{\tabcolsep}{3pt}
\begin{tabular}{@{}llcccc@{}}
\toprule
Backbone & token comparator & episodes & decision & token & GPU-h\\
\midrule
Qwen2.5-1.5B & answer only / rationale + label (same runner) & 25.6 & $9.4\pm0.8$ (2) & 2.5 / \textbf{10.0} (1) & 0.31 / 1.25--1.75\\
Qwen2.5-1.5B & GRPO (TRL), full expression & 102.4 / 38.4 & $14.0\pm0.6$ (2) & 14.3 (1) & 2.1 / 2.25\\
Qwen3.5-2B & GRPO (TRL), full expression & 102.4 / 38.4 & 47.4 (1) & 50.0 (1) & 1.7 / 2.51\\
\bottomrule
\end{tabular}
\end{table}

\paragraph{A second environment.} The same frozen comparison of the \ldm{} and the scored hybrid on FrozenLake (Qwen2.5-0.5B, history-free prompts, 60 iterations of $16\times8$ episodes, learning rate selected on validation from $\{10^{-5},10^{-4}\}$, three seeds, final checkpoints) gives $18.5\pm2.9$ vs.\ $14.6\pm1.0$\% on the 512 evaluation maps (held out by seed; 226 of 512 boards also occur among training maps) and $17.7\pm1.7$ vs.\ $15.1\pm1.2$\% on 1,024 sealed maps whose boards occur in neither training nor evaluation maps (paired intervals $[+1.6,+8.4]$ and $[+0.9,+5.9]$; with three seeds these are the smallest and largest seed differences, so they are fragile), at 0.22 vs.\ 1.22 GPU-h per run on a shared H100. The budget is short, so absolute success is low; the comparison tests the interface, not convergence.

\paragraph{Second sealed split and menu permutation.} After the first stage of the protocol (seeds 0--3), its 16 final checkpoints were evaluated once on a second sealed split (2,048 further problems, disjoint from training, test and the untouched split), with canonical menus and with one seeded permutation of every menu. Canonical: \ldm{} $44.5\pm5.2$, hybrid $41.5\pm6.8$, answer only $32.4\pm2.1$, rationale $30.0\pm1.3$\%; \ldm{} minus each: $+3.0\ [-1.4,+10.1]$, $+12.0\ [+6.9,+15.4]$, $+14.4\ [+8.4,+18.6]$. Permuted: $44.8\pm4.8$, $43.0\pm6.0$, $31.2\pm2.2$, $30.5\pm2.8$\%; $+1.8\ [-2.9,+8.8]$, $+13.5\ [+8.8,+17.1]$, $+14.3\ [+8.3,+18.6]$. The ordering, and the absence of a detected difference from the hybrid, hold on independent problems and under label reassignment.

\section{Decision-Level Case Study}
\label{app:case}
To see what the hybrid's rationale buys, we rolled out the seed-1 finalists of the \ldm{} and the scored hybrid greedily on the first 256 problems of the second sealed Countdown split, recording every decision's full label distribution (the hybrid's conditional on its generated rationale) and, with an exact solver, which successors can still reach the target (\texttt{ldmrl/case\_study.py}, \texttt{results/case\_study}). Semantically equivalent reductions are merged.

\begin{table}[H]
\caption{Decision-level comparison on 256 sealed Countdown-after problems (667 decisions per arm). Viable mass: probability on successors that can still reach the target; chosen viable: share of decisions that keep the problem solvable; $H/\log K$: normalised semantic entropy, in solved / failed episodes.}
\label{tab:case}
\centering\small
\begin{tabular}{@{}lcccccc@{}}
\toprule
Arm & solved & viable mass & chosen viable & $H/\log K$ (solved / failed) & rollout (s)\\
\midrule
\ldm{} & 108 & 0.374 & 0.421 & 0.21 / 0.74 & 11.8\\
hybrid & 107 & 0.367 & 0.421 & 0.20 / 0.64 & 77.5\\
\bottomrule
\end{tabular}
\end{table}
The two policies solve almost the same number of problems, but not the same ones (84 by both, 24 by the \ldm{} only, 23 by the hybrid only, 125 by neither). Their decisions are not detectably different: the same share of choices keeps the problem solvable, with the same probability on viable successors. The rationale therefore does not buy better decisions here; it buys time. 512 of the hybrid's 667 rationales run to the 64-token budget, and they are longer when the decision is wrong (61 tokens on average) than when it is right (35); in 145 decisions the hybrid answers almost at once (2 tokens), and these are the most reliable (viable mass 0.89). The \ldm{}'s own distribution is informative about failure: in failed episodes its normalised entropy is 0.74 against 0.21 in solved ones, and its chosen-action probability 0.40 against 0.84, so the decision plane exposes its uncertainty at no extra cost.

\input{sections/l2l_appendix}

%% file: sections/l2l_appendix.tex
\section{Decision-Plane Policies Distilled into Generative LLMs}
\label{app:l2l}

\paragraph{Protocol and supervision.}
The ALFWorld teacher is the validation-selected Qwen2.5-1.5B \ldm{} of Section~\ref{sec:distill} (\texttt{alfG\_q15\_bcrl/best}), which solves all 134 unseen and 140 seen games in 8.64 steps. It keeps its compiled memory and categorical menu readout during collection. We sample one episode at $T{=}1$ for each of 3,297 training games, excluding the 256 games held out for selection. The dump is 99.6\% successful and supplies 26,088 states from successful episodes; collection takes 1,397\,s (0.39 GPU-h) on one H100. Teacher probabilities are averaged over two shuffled menu orders before sampling.

Students are instruction-tuned LLMs with LoRA rank 16 on all language-model linear layers, starting from the base with a fresh adapter or from the teacher's \ldm{} adapter. The target is the teacher's executed command in \texttt{<action>command</action>} followed by the end-of-turn token; prompt tokens are masked, and the loss sums target-token negative log-likelihoods and averages over decisions. Two epochs give 1,630 optimiser steps at batch size 32, learning rate $5\times10^{-5}$, 5\% warmup and linear decay, AdamW with zero weight decay and gradient clipping at 1 (about 55 minutes on one H100). Evaluation uses final SFT checkpoints, a 50-step episode limit and at most 48 new tokens per action.

The prompt is verl-agent's ALFWorld template (two preceding observation/action pairs and the admissible commands); the memory prompt adds the observation-derived compiled memory that the teacher sees. The student generates each command as ordinary text; the projection lower-cases the text inside the action tags and sends it to the game unchanged.

\begin{table}[H]
\caption{Generative ALFWorld students, Qwen2.5-1.5B (\%). Selection: 256 held-out training games; unseen: 134; seen: 140. Three training seeds per row (mean, s.d.\ in brackets); $T{=}0.4$: three decoding draws of the seed-0 adapter.}
\label{tab:l2lfull}
\centering\small
\setlength{\tabcolsep}{4pt}
\begin{tabular}{@{}lrrrr@{}}
\toprule
Prompt (init.) & selection & unseen & seen & unseen, $T{=}0.4$\\
\midrule
Memory (\ldm{}) & \textbf{100.0} & \textbf{100.0} (0.0) & \textbf{100.0} (0.0) & \textbf{100.0}\\
Memory (base), seed 0 & 99.2 & \textbf{100.0} & \textbf{100.0} & \textbf{100.0}\\
No memory (base) & 96.2 & 89.3 (3.4) & 95.7 & 90.5\\
\midrule
Published GiGPO & & 86.7 (1.7) & & 86.7\\
\bottomrule
\end{tabular}
\end{table}

The three memory-prompt seeds solve every selection, unseen and seen game in 8.4 steps per unseen game (teacher: 8.64), with no malformed or non-admissible output; each $T{=}0.4$ draw of both seed-0 adapters also solves every unseen and seen game. Without memory the three $T{=}0.4$ draws give 89.6, 91.8 and 90.3\% (game-bootstrap interval 84.3--93.8), against GiGPO's 86.7\%. For reference, the base model solves 2.2\% of unseen games zero-shot.

\paragraph{Measured end-to-end cost.}
Table~\ref{tab:l2lcost} times the complete route as one run on an exclusive H100, from a Qwen3.5-2B \ldm{} trained from scratch to the evaluated generative Qwen2.5-1.5B, and compares it with token-level GiGPO of the same LLM in the same runner, with the same memory prompt and LoRA capacity. The route reaches 99.3\% unseen (seen 99.3, selection 98.0) after 12.70 GPU-h; token RL reaches 86.6\% (selection 88.7) after 17.34 GPU-h and 150 iterations, and 83.6\% at 15.8 GPU-h. Published GiGPO's 362.83\,s per iteration on two H100s for 150 iterations amounts to about 30.3 GPU-h.

\begin{table}[H]
\caption{Allocated H100 GPU-hours of the complete route (one exclusive run) and of token-level RL of the same LLM.}
\label{tab:l2lcost}
\centering\small
\setlength{\tabcolsep}{4pt}
\begin{tabular}{@{}lrr@{}}
\toprule
Stage & GPU-h & unseen\\
\midrule
Qwen3.5-2B \ldm{}-RL from scratch, 150 it & 9.21 & \\
Teacher dump and BC into the 1.5B \ldm{} & 0.62 & \\
1.5B \ldm{}-RL, 100 it & 2.29 & \\
Rollouts, 100-step SFT, evaluation & 0.59 & \\
\textbf{\rlid{} route, total} & \textbf{12.70} & \textbf{99.3}\\
\midrule
Token-level GiGPO, same LLM and runner, 150 it & 17.34 & 86.6\\
\bottomrule
\end{tabular}
\end{table}

\paragraph{Cheap conversion.}
Given a trained 1.5B \ldm{}, collection (1,397\,s) and a 100-step SFT run (280\,s, 4.7 minutes) cost 0.47 GPU-h and give 100\% unseen success. The \ldm{} adapter is an excellent initialisation for action generation: it gives 97.8\% unseen after only 25 SFT steps (Table~\ref{tab:l2lbudget}). Each budget is a separate run with its own warmup/decay schedule.

\begin{table}[H]
\caption{Separate SFT-budget runs: ALFWorld unseen success (\%, memory prompt, greedy).}
\label{tab:l2lbudget}
\centering\small
\begin{tabular}{@{}lrrrr@{}}
\toprule
Initialisation & 25 steps & 50 steps & 100 steps & 200 steps\\
\midrule
Base & 15.7 & 26.9 & 67.9 & --\\
\ldm{} adapter & \textbf{97.8} & \textbf{97.8} & \textbf{100.0} & \textbf{100.0}\\
\bottomrule
\end{tabular}
\end{table}

\paragraph{Deployment without an action list.}
A student whose prompt omits the admissible commands, while the teacher keeps acting on its menu, reaches 99.3\% unseen and 100\% seen from base initialisation (selection 98.4), and 99.3\% / 99.3\% when warm-started from the memory-prompt student (non-admissible commands 0.9\% of unseen decisions). Three DAgger rounds on this student lower non-admissible commands on selection games from 6.2\% to 1.5\% and raise selection success from 98.8\% to 99.2\%.

\paragraph{Across model families and sizes.}
With the same 26,088 teacher states and the memory prompt, Granite-3.3-2B-Instruct solves every selection, unseen and seen game with no non-admissible output; SmolLM2-1.7B-Instruct reaches 96.1 / 96.3 / 99.3\% and Qwen2.5-0.5B-Instruct 98.4 / 98.5 / 100\% (selection / unseen / seen). Granite's loss applies its $\div 8$ logit scaling.

\paragraph{Which decision-plane teacher?}
Distilling directly from the final Qwen3.5-2B \ldm{} (98.5\% greedy) with the memory prompt gives 99.3\% unseen, 100\% seen and 98.8\% selection in 9.25 steps, for about 1.4 GPU-h after the teacher (0.52 collection, 0.92 SFT). Without memory in the student prompt, the memory-trained 1.5B teacher gives 91.8\% unseen, well above the 78.4\% from a Qwen3.5-2B teacher trained without memory (Table~\ref{tab:l2lbypass}).

\begin{table}[H]
\caption{Teacher choice. Students start from Qwen2.5-1.5B base weights; final two-epoch SFT, greedy ALFWorld success (\%).}
\label{tab:l2lbypass}
\centering\small
\begin{tabular}{@{}llrr@{}}
\toprule
Teacher & student prompt & unseen & seen\\
\midrule
Qwen3.5-2B \ldm{}, with memory & memory & 99.3 & \textbf{100.0}\\
Refined 1.5B \ldm{}, with memory & memory & \textbf{100.0} & \textbf{100.0}\\
Qwen3.5-2B \ldm{}, trained without memory & no memory & 78.4 & 82.9\\
Refined 1.5B \ldm{}, with memory & no memory & \textbf{91.8} & \textbf{96.4}\\
\bottomrule
\end{tabular}
\end{table}

\paragraph{The \ldm{} as a verifier.}
At inference the \ldm{} can also rerank proposals: the generative student samples four commands at $T{=}1$ and the \ldm{} picks the admissible one it rates highest. For SmolLM2-1.7B this lifts unseen success from 96.3\% to 100\% and selection success from 96.1\% to 99.6\% (Table~\ref{tab:l2lrerank}). Paired by game, \ldm{} selection beats taking the first admissible proposal 8--0 on unseen games (exact McNemar $p=0.0078$) and 9--1 on selection games ($p=0.021$). It also lifts the menu-free student's selection success from 98.8\% to 100\%.

\begin{table}[H]
\caption{Choosing one of four sampled proposals, SmolLM2-1.7B student (\%).}
\label{tab:l2lrerank}
\centering\small
\begin{tabular}{@{}lrr@{}}
\toprule
Selector & selection & unseen\\
\midrule
None (greedy, one sample) & 96.1 & 96.3\\
First admissible proposal & 96.5 & 94.0\\
Most frequent admissible proposal & 98.0 & 97.8\\
\ldm{}'s highest-probability proposal & \textbf{99.6} & \textbf{100.0}\\
\bottomrule
\end{tabular}
\end{table}

\paragraph{A teacher-preference bonus speeds up token RL.}
The \ldm{} can also guide token-level RL of the base LLM. Each decision's GiGPO-normalised advantage receives a bonus $u=\mathrm{clip}(\log p(a)+H(p),-2,2)$, where $p$ is the \ldm{}'s distribution over the menu commands ($u=-2$ for malformed or inadmissible commands), with weight 0.2 annealed to 0 over 40 iterations. After 30 iterations the bonus runs reach 26.6\% and 20.7\% selection success, against 10.9\% and 0.4\% for plain GiGPO with the same seeds (Table~\ref{tab:l2lbonus}).

\begin{table}[H]
\caption{Greedy selection success (\%, 256 games) of token-level RL from the base model, with and without the \ldm{} preference bonus.}
\label{tab:l2lbonus}
\centering\small
\begin{tabular}{@{}lrrrr@{}}
\toprule
Run & it 10 & it 20 & it 30 & GPU-h at 30\\
\midrule
Bonus, seed 0 & 2.7 & 9.8 & \textbf{26.6} & 4.81\\
Bonus, seed 1 & 3.1 & 8.2 & \textbf{20.7} & 5.11\\
GiGPO, seed 0 & 2.7 & 5.1 & 10.9 & 4.16\\
GiGPO, seed 1 & 0.8 & 0.0 & 0.4 & 4.15\\
\bottomrule
\end{tabular}
\end{table}

\paragraph{Why token RL is slow: staying on the list.}
Token-level RL of the same LLM learns the output format within about four iterations (0.6 GPU-h), then spends most of its budget learning to stay on the admissible list (Table~\ref{tab:l2lwhy}). Each off-list command wastes a step, episodes run to the 50-step limit, and GiGPO groups with equal returns carry no signal. In diagnostic rollouts the base model's outputs are 96\% malformed (untagged text, wrong tags or unclosed tags); after 30 iterations the format is learned, but 71\% of commands are off the list, mostly repeats of a command that is no longer admissible (54\%) and empty actions (29\%). The \ldm{} outputs a softmax over listed labels, so every sample is a legal action from the first iteration. With a Qwen3.5-2B backbone, which follows the format and stays mostly on the list, the \ldm{} still reaches 76.1\% at iteration 40 in 3.3 GPU-h, against 19.9\% for token RL in 7.0 GPU-h.

\begin{table}[H]
\caption{Token-level RL of Qwen2.5-1.5B from the base model (memory prompt): per-decision rates at $T{=}1$ (\%).}
\label{tab:l2lwhy}
\centering\small
\begin{tabular}{@{}rrrrr@{}}
\toprule
Iteration & train success & malformed & non-admissible & episode length\\
\midrule
1 & 0.0 & 94.7 & 91.2 & 50.0\\
4 & 1.6 & 0.6 & 72.4 & 49.4\\
10 & 7.8 & 0.0 & 53.4 & 46.8\\
30 & 28.9 & 0.0 & 38.8 & 39.3\\
70 & 73.4 & 0.0 & 12.0 & 24.7\\
150 & 91.4 & 0.0 & 7.4 & 14.4\\
\midrule
\ldm{}, any iteration & & 0.0 & 0.0 & \\
\bottomrule
\end{tabular}
\end{table}

\paragraph{General ability.}
The distilled students keep the base model's general ability on the first 500 GSM8K test questions and all 1,172 ARC-Challenge test questions; the no-memory student scores above the base on both (Table~\ref{tab:l2lgeneral}). GSM8K uses zero-shot chain-of-thought prompting, a boxed numerical answer and at most 512 generated tokens; ARC asks for a letter read from free generation, at most 16 tokens. Both are greedy.

\begin{table}[H]
\caption{General-ability checks (\%) with the SFT adapter enabled.}
\label{tab:l2lgeneral}
\centering\small
\begin{tabular}{@{}lrr@{}}
\toprule
Policy & GSM8K (500) & ARC-C (1,172)\\
\midrule
Base & 74.6 & 75.3\\
Memory prompt, \ldm{} init. & 73.8 & 75.1\\
Memory prompt, base init. & 73.4 & 75.6\\
No memory, base init. & \textbf{75.8} & \textbf{75.8}\\
\bottomrule
\end{tabular}
\end{table}

\paragraph{Sokoban.}
The teacher is \texttt{sokL\_q05\_s2/best}, a Qwen2.5-0.5B \ldm{} (82.8\% on the 512 evaluation maps, 35.9\% on the 512 OOD maps). We collect 8,000 episodes at $T{=}1$ on training seeds, excluding the 256 selection puzzles; the 6,567 successful episodes supply 26,451 states, collected in about three minutes on one H100. Two-epoch SFT takes 1,653 steps and about 12 minutes. The generative students match and exceed the teacher in distribution (Table~\ref{tab:l2lsok}).

\begin{table}[H]
\caption{Sokoban teacher and generative students (\%, greedy). Eval and OOD each contain 512 maps.}
\label{tab:l2lsok}
\centering\small
\begin{tabular}{@{}lrr@{}}
\toprule
Policy & eval & OOD\\
\midrule
\ldm{} teacher & 82.8 & \textbf{35.9}\\
Generative SFT, base init. & 83.2 & 29.5\\
Generative SFT, \ldm{} init. & \textbf{83.4} & 34.4\\
\bottomrule
\end{tabular}
\end{table}

\begingroup\raggedright
\paragraph{Artifacts.}
Final evaluations, paired outcomes, sampled reads and SFT timings are archived in \texttt{llmdistill/raw/}; cross-family students, verifier runs, allocation clocks and token-RL metrics in \texttt{llmdistill/raw/track2/}; diagnostic rollouts in \texttt{llmdistill/raw/why/}. Prompt rendering, projection and collection are implemented in \texttt{ldmrl/l2l.py}, the SFT loss and schedules in \texttt{ldmrl/l2l\_sft.py}, token-level GiGPO in \texttt{ldmrl/l2l\_rl.py} and the general checks in \texttt{ldmrl/l2l\_general.py}.
\par\endgroup